\documentclass{article}

\usepackage{PRIMEarxiv}

\usepackage{microtype}
\usepackage{graphicx}
\usepackage{subfigure}
\usepackage{booktabs}
\usepackage[utf8]{inputenc} 
\usepackage[T1]{fontenc}    
\usepackage{hyperref}       
\usepackage{url}            
\usepackage{booktabs}       
\usepackage{amsfonts}       
\usepackage{nicefrac}       
\usepackage{microtype}      
\usepackage{lipsum}
\usepackage{fancyhdr}       
\usepackage{graphicx}       
\graphicspath{{media/}}     
\usepackage{amsmath}
\usepackage{amssymb}
\usepackage{mathtools}
\usepackage{amsthm}

\usepackage{bbm, dsfont}
\theoremstyle{plain}
\newtheorem{theorem}{Theorem}[section]

\newtheorem{corollary}[theorem]{Corollary}
\theoremstyle{definition}

\newtheorem{assumption}[theorem]{Assumption}
\theoremstyle{remark}
\newtheorem{remark}[theorem]{Remark}
\DeclareMathOperator*{\argmax}{arg\,max}
\DeclareMathOperator*{\argmin}{arg\,min}
\usepackage{natbib}

\title{To be Robust and to be Fair: Aligning Fairness with Robustness}

\author{
  Junyi Chai, Xiaoqian Wang \\
  Elmore Family School of Electrical and Computer Engineering\\
  Purdue University \\
  \texttt{\{chai28,joywang\}@purdue.edu} \\}

\begin{document}
\maketitle

\begin{abstract}
Adversarial training has been shown to be reliable in improving robustness against adversarial samples. However, the problem of adversarial training in terms of fairness has not yet been properly studied, and the relationship between fairness and accuracy attack still remains unclear. Can we simultaneously improve robustness w.r.t. both fairness and accuracy? To tackle this topic, in this paper, we study the problem of adversarial training and adversarial attack w.r.t. both metrics. We propose a unified structure for fairness attack which brings together common notions in group fairness, and we theoretically prove the equivalence of fairness attack against different notions. Moreover, we show the alignment of fairness and accuracy attack, and theoretically demonstrate that robustness w.r.t. one metric benefits from robustness w.r.t. the other metric. Our study suggests a novel way to unify adversarial training and attack w.r.t. fairness and accuracy, and experimental results show that our proposed method achieves better performance in terms of robustness w.r.t. both metrics.
\end{abstract}

\section{Introduction}

As machine learning systems have been increasingly applied in social fields, it is imperative that machine learning models do not reflect real-world discrimination. However, machine learning models have shown biased predictions against disadvantaged groups on several real-world tasks \citep{larson2016compas,dressel2018accuracy,mehrabi2021survey}. In order to improve fairness and reduce discrimination of machine learning systems, a variety of work has been proposed to quantify and rectify bias \citep{hardt2016equality,kleinberg2016inherent,mitchell2018prediction}.

\begin{figure}[!h]
    \centering
    \includegraphics[width=1\linewidth,height=0.6\linewidth]{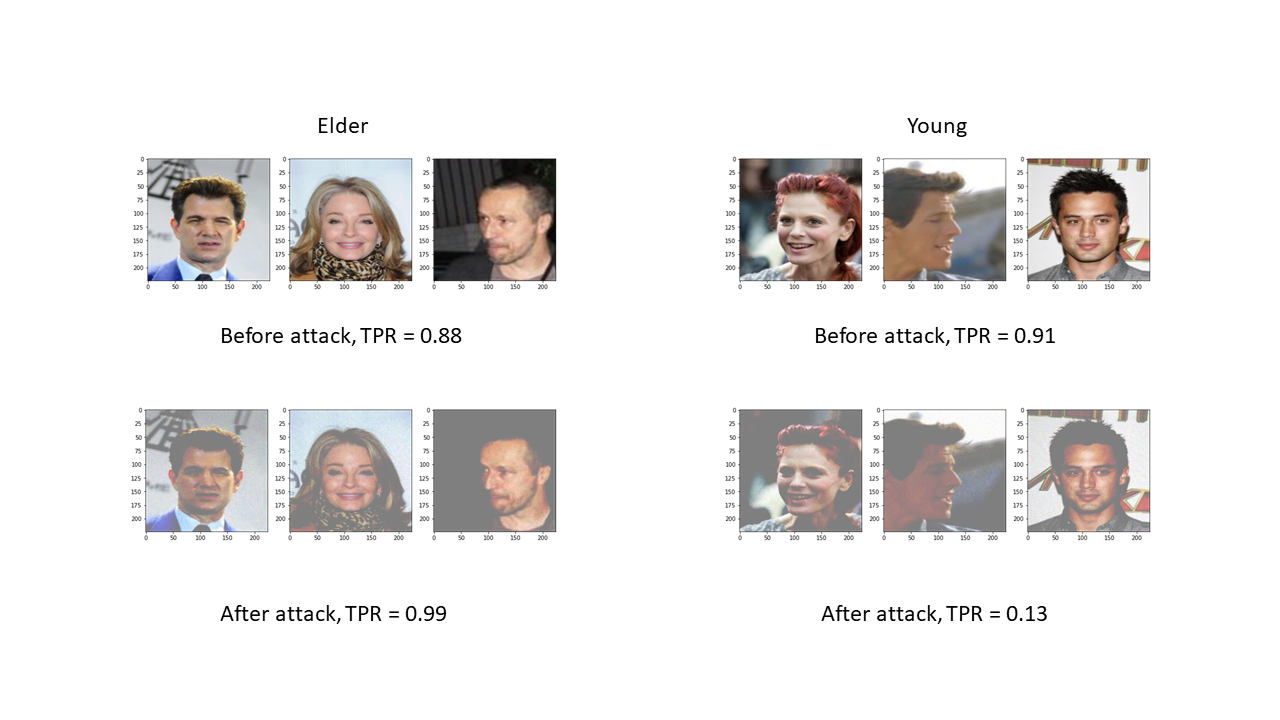}
    \vspace{-20pt}
    \caption{Demonstration of adversarial attacks against fairness on CelebA dataset. Even under small perturbation level $\epsilon=0.1$, the disparities in true positive rate (TPR) between young and old elder people increase sharply from $0.03$ to $0.86$. This shows that fairness shall not be considered as a static measure, and small fairness gaps can lead to large disparities under adversarial attacks. Enforcing fairness alone during training does not necessarily lead to robustness against fairness attacks, and it is important to consider the robustness of fair classifiers during training. }
    \label{demo}
\end{figure}

Despite the emerging interest in fairness, the topic of adversarial fairness attack and robustness against such attack have not yet been properly discussed. Most of current literature on adversarial training has been focusing on improving robustness against accuracy attack \citep{chakraborty2018adversarial}, while the problem of adversarial attack and adversarial training w.r.t. fairness has been rarely addressed. However, as with accuracy, fairness can also be targeted by adversarials, where attacks against fairness can be employed to depreciate trustworthiness of models, or can be intentionally employed by individuals who gain profits from the biased decisions against certain groups. Besides, fairness can be volatile under adversarial perturbations, where a small degree of perturbation can lead to significant variations in group-wise disparities, as shown in Fig. \ref{demo}. Therefore, enforcing fairness alone during training does not necessarily leads to improvement in fairness robustness, and robustness against fairness attack is non-trivial and worth consideration when developing models.

{Although adversarial learning has been widely discussed in fairness literature,} much of them have been focusing on applying adversarial learning as means to unlearn the impact of sensitive attributes to achieve fairness \citep{madras2018learning,creager2019flexibly}.
\citet{solans2020poisoning} and \citet{mehrabi2021exacerbating} made the first attempt to propose novel ways to generate adversarial samples taking into account fairness objectives to disturb the training process and exacerbate bias on clean testing data. However, 
{it remains unclear how to improve robustness against fairness attack,} and the relationship between fairness and accuracy attack is understudied.
In light of current limitations on adversarial attack and adversarial training w.r.t. fairness, in this work, we propose a general framework for fairness attack, where we show impacts of fairness attacks up to each individual under different notions, and the connections between these notions regarding gradient-based attacks. Based on this unified framework, we discuss the relationship between fairness and accuracy attack, and we show theoretically how robustness w.r.t. fairness and accuracy can benefit from each other, i.e., the alignment between the two notions in terms of robustness. Our discussion suggests a novel framework, \textit{fair adversarial training}, to incorporate fair classification with adversarial training to improve robustness against fairness attack. We summarize our contribution as follows:
\begin{itemize}
    \item We propose a unified framework for adversarial attack in fairness, {which brings together different notions in group fairness.}
    \item We theoretically demonstrate the alignment between robustness of fairness and accuracy, and we propose a \textit{fair adversarial training} approach that incorporates adversarial training with fair classification.
    \item We empirically validate the superiority of our method under adversarial attack, and the connection between robustness w.r.t. fairness and accuracy on four benchmark datasets.
\end{itemize}

\section{Related Work}

\subsection{Fairness in machine learning}

Fairness has gained much attention in machine learning society. Different notions have been proposed to quantify discrimination of machine learning models, including individual fairness \citep{lahoti2019operationalizing,john2020verifying,mukherjee2020two}, group fairness \citep{feldman2015certifying,hardt2016equality,zafar2017fairness} and counterfactual fairness \citep{kusner2017counterfactual}. Our work is most closely related with group fairness notions. Works on group fairness generally fall into three categories: preprocessing \citep{creager2019flexibly,jiang2020identifying,jang2021constructing}, where the goal is to adjust training distribution to reduce discrimination; inprocessing \citep{zafar2017fairness,jung2021fair,roh2021sample}, where the goal is to impose fairness constraint during training by reweighing or adding relaxed fairness regularization; and postprocessing \citep{hardt2016equality,jang2022group}, where the goal is to adjust the decision threshold in each sensitive group to achieve expected fairness parity.

\subsection{Adversarial machine learning}
Adversarial training and adversarial attack have been widely studied in trustworthy machine learning. \citet{goodfellow2014explaining} propose a simple one-step gradient-based attack to adversarially perturb the predicted label. \citet{madry2017towards} extend the one-step attack to an iterative attack strategy, and show that iterative strategy is better at finding adversarial samples. Accordingly, different methods on adversarial attack and defense have been proposed \citep{shafahi2019adversarial,wong2020fast,xie2020smooth,cui2021learnable,jia2022boosting} to improve robustness of classifier against accuracy attack. However, few literature has addressed adversarial training and attack against fairness. Some work discusses the problem of fairness poisoning attack during training \citep{solans2020poisoning,mehrabi2021exacerbating}; however, it is not clear how fairness attack would influence the predicted soft labels, and the relationship between fairness and accuracy attack/robustness remains unclear.

\section{Problem Definition}

\subsection{Adversarial attack w.r.t. accuracy}
We start by formulating adversarial attack against accuracy. 
Denote $x \in \mathbb{R}^n$ as the input feature, $y \in \{0,1\}$ as the label, and $a \in \{0,1\}$ as the 
sensitive attribute\footnote{We formulate the problem under binary classification and binary sensitive attribute for simplicity; however, our method can be readily generalized to multi-class and multi-sensitive-attribute scenarios.}.
Let $f: \mathbb{R}^n\rightarrow[0,1]$ be the function of classifier, the objective of adversarial attack against accuracy for each sample $(x, y, a)$ can be formulated as
\begin{equation*}
\argmax_{\epsilon} L_\text{CE}(f(x+\epsilon),y),~\text{s.t.} \|\epsilon\| \le \epsilon_0,
\end{equation*}
where $\|\epsilon\|$ refers to the $L^p$ norm of $\epsilon$ with a general choice of perturbation constraint in $L^{\infty}$ norm, and $L_\text{CE}$ is the cross-entropy loss. A common way to obtain adversarial samples is through projected gradient descent (PGD) attack, where the adversarial sample is updated in each step based on the signed gradient:
\begin{equation*}
x^{t+1}=\Pi_{x+S}\left(x^t+\alpha \operatorname{sign}\left(\nabla_x L_\text{CE}(x, y)\right)\right),
\end{equation*}
where $\alpha$ is the step size, and $S := \{\epsilon, \|\epsilon\| \le \epsilon_0\}$ is the set of allowed perturbation. PGD attack has been shown to be effective in finding adversarial samples compared with one-step adversarial attack \citep{madry2017towards}.
\subsection{Adversarial attack w.r.t. fairness}
Fairness adversarial attack has yet been less studied in current literature. In light of the formulation of accuracy adversarial attack, we propose to formulate fairness adversarial attack as follows:

\begin{equation*}
\argmax_{\epsilon} L(f(x+\epsilon),y),~\text{s.t.} \|\epsilon\| \le \epsilon_0,
\end{equation*}
where $L$ is some relaxed fairness constraint. Below we discuss the specific formulation of $L$ for two widely adopted group fairness notions: disparate impact (DI) and equalized odds (EOd).

Consider a testing set $\mathbb{S} = \{(x_i,y_i,a_i),1\le i \le N\}$ and denote $\mathbb{S}_{jk} = \{x_i | y_i = j, a_i = k\}$, and $\mathbb{S}_{.k} = \{x_i | a_i = k\}$.
The relaxations \citep{madras2018learning,wang2022understanding} for fairness attack corresponding to DI and EOd can be formulated as:
\begin{equation}
\label{eq:DP}
L_{DI} = \left| \sum_{x_i \in \mathbb{S}_{.1}} \frac{f(x_i)}{|\mathbb{S}_{.1}|} - \sum_{x_i \in \mathbb{S}_{.0}}\frac{f(x_i)}{|\mathbb{S}_{.0}|}\right|,
\end{equation}
\begin{equation}
 \label{eq:EOd}
 L_{EOd} = \sum_{y} \left|\sum_{x_i \in \mathbb{S}_{y0}}\frac{f(x_i)}{|\mathbb{S}_{y0}|}-\sum_{x_i \in \mathbb{S}_{y1}} \frac{f(x_i)}{|\mathbb{S}_{y1}|}\right|,
 \end{equation}

And fair adversarial samples can be obtained similarly via PGD attack:
\begin{equation*}
x^{t+1}=\Pi_{x+S}\left(x^t+\alpha \operatorname{sign}\left(\nabla_x L(x, y)\right)\right).
\end{equation*}

\section{Aligning Fairness with Robustness}


{We will discuss in this section the alignment of fairness with robustness in three folds: 1) connection between attacks in different fairnesss notions; 2) connection between fairness and accuracy attack; 3) connection between robustness against fairness attack and that against accuracy attack.}

Before going to the discussion, we first clarify the notations. 
We denote as $x_{\text{sub},a}^{t, \text{obj}}$ the adversarial sample(s) of the specified subgroup (sub) in sensitive group $a$ at $t$-th iteration under the attack designated by obj. For example, $x_{\text{TP},0}^{t, \text{Acc}}$ refers to the adversarial true positive (TP) sample(s) in the disadvantaged group ($a=0$) at $t$-th iteration under accuracy attack. {We will slightly abuse the notation $x_{\text{sub},a}^{t, \text{obj}}$ in this section to denote both one individual sample and the set of samples in the subgroup.}

Without loss of generality, in this section we assume the positive label is the favorable outcome for classification, and we assume $a=1$ is the advantaged group\footnote{Here we define advantaged group as the one with higher average positive prediction.}.

\subsection{Connection between EOd and DI attack}
We now discuss the detailed relationship between DI and EOd attack. The following corollary states the compatibility of the two objectives: 
\begin{corollary}
\label{cor_1}
The adversarial objective of EOd attack is lower-bounded by that of DI attack up to multiplicative constants.
\end{corollary}
We defer the proof to appendix. Corollary \ref{cor_1} shows the connection between adversarial attack against different group fairness notions, where attack targets at group fairness perturb the predicted soft labels against sensitive attributes. Specifically, a successful DI attack also leads to a successful EOd attack, while the opposite does not necessarily hold true. In light of this, we will focus on DI attack for the rest of this paper.
 
For $(x_j, y_j, a_j)$ in the advantaged group, we can rewrite $L_{DI}$ in \eqref{eq:DP} as:
\begin{equation}
\begin{aligned}
L_{DI} & = \left| \sum_{x_i \in \mathbb{S}_{.1}} \frac{f(x_i)}{|\mathbb{S}_{.1}|} - \sum_{x_i \in \mathbb{S}_{.0}}\frac{f(x_i)}{|\mathbb{S}_{.0}|}\right|
\\
& = \frac{f(x_j)}{|\mathbb{S}_{.a_j}|} + \sum_{x_i \in \mathbb{S}_{.a_j}\backslash\{x_j\}} \frac{f(x_i)}{|\mathbb{S}_{.a_j}|} - \sum_{x_i \in \mathbb{S}_{.\hat{a}_j}}\frac{f(x_i)}{|\mathbb{S}_{.\hat{a}_j}|}
\\
& = \frac{f(x_j)}{|\mathbb{S}_{.a_j}|} + C_j,
\end{aligned}
\label{adv_eod}
\end{equation}
where {$\hat{a}_j=\left|1-a_j\right|$} and $C_j$ is a constant w.r.t. $x_j$ {thus does not affect $\frac{\partial L_{DI}}{\partial x_j}$.} This shows that the DI attack is expected to maximize the prediction in the advantaged group.

Similarly, for sample $(x_k, y_k, a_k)$ in the disadvantaged group, we have:
\begin{equation}
\begin{aligned}
L_{DI} & = \left| \sum_{x_i \in \mathbb{S}_{.1}} \frac{f(x_i)}{|\mathbb{S}_{.1}|} - \sum_{x_i \in \mathbb{S}_{.0}}\frac{f(x_i)}{|\mathbb{S}_{.0}|}\right|
\\
& = {-\frac{f(x_k)}{|\mathbb{S}_{.a_k}|} + C_k,}
\end{aligned}
\label{dis_eod}
\end{equation}
where $C_k$ is a constant w.r.t. $x_k$ {thus does not affect $\frac{\partial L_{DI}}{\partial x_k}$.} \eqref{dis_eod} shows that the DI attack in disadvantaged group is contrary to that of advantaged group, where the predictions are expected to be \emph{minimized}.
\subsection{Alignment between DI and accuracy attack} \label{ali_acr_eod}

We move on to discuss the connection between DI and accuracy attack. 
The following corollary shows the connection between fairness and accuracy attack:
\begin{corollary}
The fairness adversarial attack and accuracy adversarial attack are in alignment regarding true negative (TN) and false positive (FP) samples in advantaged group and true positive (TP) and false negative (FN) samples in disadvantaged group. 
\label{cor_2}
\end{corollary}

We defer the detailed proof to appendix. {It is worth noticing that the fairness adversarial attack and accuracy adversarial attack operates towards the opposite direction for the the rest of samples.} 
That is, for the two subgroups $x_{\text{TP},1}$ and $x_{\text{TN},0}$: as in \eqref{adv_eod} and \eqref{dis_eod}, the adversarial attack regarding fairness is expected to {maximize their predicted confidence} (i.e., maximizing the predicted soft labels for $x_{\text{TP},1}$, and minimizing the predicted soft labels for $x_{\text{TN},0}$), 
{while adversarial attack regarding accuracy is expected to minimize the predicted confidence for these two subgroups.} 

Similarly, for the two subgroups $x_{\text{FN},1}$ and $x_{\text{FP},0}$: the fairness attack is expected to `correct' their predicted soft labels, i.e., the predicted adversarial labels are expected to be in alignment with ground-truth labels, while accuracy attack is expected to exacerbate their error. We summarize the connection between fairness and accuracy attack on various subgroups in Table \ref{tb:alignment}.

\begin{table}[!h]
\centering
\begin{tabular}{lll}
\hline
Sensitive Group & Alignment & Misalignment \\
\hline
Disadvantaged ($a=0$) & $x_{\text{TP},0}$, $x_{\text{FN},0}$ & $x_{\text{TN},0}$, $x_{\text{FP},0}$ \\
\hline
Advantaged ($a=1$) & $x_{\text{TN},1}$, $x_{\text{FP},1}$ & $x_{\text{TP},1}$, $x_{\text{FN},1}$ \\
\hline
\end{tabular}
\caption{{Connection between fairness and accuracy attack regarding samples in different subgroups.} Alignment/Misalignment means fairness attack and accuracy attack operate in the same/opposite directions in the subgroups, respectively.}
\label{tb:alignment}
\end{table}

\subsection{Alignment between fairness and accuracy robustness}

We now discuss the alignment between robustness w.r.t. fairness and accuracy.
According to Table \ref{tb:alignment}, the relationship between robustness w.r.t. fairness and accuracy is straightforward on the four subgroups in `Alignment' category, since the fairness attack and accuracy attack operate in the same direction for those samples. However, the relationship on the four subgroups in `Misalignment' category is not clear.

Therefore, in the following we focus our discussion on the four misalignment subgroups in Table \ref{tb:alignment}: $x_{\text{TP},1}$, $x_{\text{FN},1}$, $x_{\text{TN},0}$, $x_{\text{FP},0}$. Before we state the detailed relationship, we first state the assumption we need to prove the relationship:

\begin{assumption}
The gradient of $f$ w.r.t. input feature $x$ is Lipschitz with constant $K$.
\label{lip}
\end{assumption}

Under Assumption \ref{lip}, below we discuss the alignment between fairness and accuracy robustness in two directions.

\subsubsection{From accuracy robustness to fairness robustness}
we first derive the guarantee for robustness against DI attack by robustness under accuracy attack. We will focus on $x_{\text{FN},1}$ and $x_{\text{FP},0}$, {as DI attack regarding $x_{\text{TP},1}$ and $x_{\text{TN},0}$ does not affect fairness.}

\begin{theorem}
Given a classifier $f$, consider $\epsilon$-level DI attack with step size $\alpha$ and up to $T$ iterations, let $D(x):=|L(f({x^\text{DI}}),y)-L(f(x),y)|$ be the change of cross-entropy loss for sample $x$ under DI attack, 
{let $\delta_{\text{FN},0}$ be the change of $f(x_{\text{FN}, 0})$ under $\epsilon$-level accuracy attack,} 
then the difference of fairness robustness between $x_{\text{FN},1}$ and $x_{\text{FN},0}$ is upper-bounded by the accuracy robustness of $x_{\text{FN},0}$ up to an addictive and a multiplicative constant: 
\begin{equation*}
{|D(x_{\text{FN}, 1})-D(x_{\text{FN}, 0})|} \leq \min_{x_{\text{FN}, 0} \in \mathbb{S}_{10}} \alpha\sum_{t=1}^{T} G_t,
\end{equation*}
\begin{equation*}
G_t=\left[\frac{\sqrt{n} Kd(x_{\text{FN}, 1}^{t-1,\text{DI}},x_{\text{FN}, 0}^{t-1,\text{DI}})}{f(x_{\text{FN}, 1}^{t-1,\text{DI}})} +  \eta_t \delta_{\text{FN},0}^{t-1}\right],
\end{equation*}
\begin{equation*}
\eta_t = \left|\frac{ f(x_{\text{FN}, 0}^{t-1,\text{DI}}) - f(x_{\text{FN}, 1}^{t-1,\text{DI}}) }{f(x_{\text{FN}, 1}^{t-1,\text{DI}})f(x_{\text{FN}, 0}^{t-1,\text{DI}})} \right|.
\end{equation*}
\label{acr-eod}
\end{theorem}
Detailed proof can be found in the appendix. As discussed in Section \ref{ali_acr_eod}, fairness robustness of $x_{\text{FN}, 0}$ benefits directly from adversarial training w.r.t. accuracy, 
{while robustness of $x_{\text{FN}, 1}$ are not clearly related with adversarial training w.r.t. accuracy. Instead, we compare with $x_{\text{FN}, 0}$ to provide robustness guarantee for $x_{\text{FN}, 1}$ samples against DI attack.} 
{Specifically, for $f'$ under accuracy adversarial training and $f$ under normal training, we have similar upper-bound, except that {we now have $\delta_{\text{FN},0}^{'t-1} \le \delta_{\text{FN},0}^{t-1}$, which indicates a tighter upper-bound for $f'$ in Theorem \ref{acr-eod}.} }
For {marginal advantaged FN samples ($x_{\text{FN}, 1}$) which are more vulnerable under DI attack, we have their robustness bounded by marginal disadvantaged FN samples ($x_{\text{FN}, 0}$), and smaller $\delta_{\text{FN},0}$, or tighter bound indicates better robustness.} In this way, classifiers under accuracy adversarial training also achieve improvement in fairness robustness. Similar inequality in Theorem \ref{acr-eod} also holds for $x_{\text{FP},0}$ and $x_{\text{FP},1}$:
\begin{remark}
For $x_{\text{FP},0}$ and $x_{\text{FP},1}$, we have similar inequality regarding the upper-bound of robustness difference:
$$
|D(x_{\text{FP},0})-D(x_{\text{FP},1})| \leq \min_{x_{\text{FP},1} \in \mathbb{S}_{01}} \alpha\sum_{t=1}^{T} {H_t}, 
$$
$$
H_t=\left[\frac{\sqrt{n} Kd(x_{\text{FP}, 0}^{t-1,\text{DI}},x_{\text{FP}, 1}^{t-1,\text{DI}})}{{ f(x_{\text{FP}, 0}^{t-1,\text{DI}})}} +  \rho_t {\delta_{\text{FP},1}^{t-1}}\right],
$$
$$
\rho_t = {\left|\frac{ f(x_{\text{FP}, 0}^{t-1,\text{DI}}) - f(x_{\text{FP}, 1}^{t-1,\text{DI}}) }{f(x_{\text{FP}, 1}^{t-1,\text{DI}})f(x_{\text{FP}, 0}^{t-1,\text{DI}})} \right|.}
$$
\label{rm: acr-eod2}
\end{remark}

\subsubsection{From fairness robustness to accuracy robustness}
For the reversed direction, under Assumption \ref{lip}, we have the following guarantee for robustness against accuracy attack by robustness against DI attack. We will focus on $x_{\text{TP},1}$ and $x_{\text{TN},0}$, {as accuracy attack regarding $x_{\text{FN},1}$ and $x_{\text{FP},0}$ does not affect accuracy.}
\begin{theorem}
Given a classifier $f$, consider $\epsilon$-level accuracy attack with step size $\alpha$ and up to $T$ iterations, let $F(x) := |f(x^\text{Acc})-f(x)|$ be the change of predicted soft label under $\epsilon$-level accuracy attack, let {$\xi_{\text{TP},0}$} be the change of $f(x_{\text{TP},0})$ under $\epsilon$-level DI attack, then the accuracy robustness of $x_{\text{TP},1}$ is {upper-bounded} by the fairness robustness of $x_{\text{TP},0}$ up to an addictive constant: 
\begin{equation*}
F(x_{\text{TP},1}) \le \min_{x_{\text{TP},0} \in \mathbb{S}_{10}} \xi_{\text{TP},0} +\sum_{t=1}^T \sqrt{n}\alpha Kd(x^{t-1, \text{Acc}}_{\text{TP},1},x^{t-1,\text{Acc}}_{\text{TP},0}).
\end{equation*}
\label{eod-acr}
\end{theorem}
Here the adversarial attack against fairness and accuracy are in alignment regarding $x_{\text{TP},0}$, which we use to upper-bound robustness of $x_{\text{TP},1}$ under accuracy attack. Theorem \ref{eod-acr} shows that adversarial training w.r.t. fairness also benefits robustness w.r.t. accuracy. 
{Specifically, for $f''$ under fairness adversarial training and $f$ under normal training, we have similar inequality, except that {we now have $\xi''_{\text{TP},0}\le \xi_{\text{TP},0}$,} which indicates better robustness for $x_{\text{TP},1}$ under adversarial training.}
Similar upper-bound also holds for TN samples:

\begin{remark}
For $x_{\text{TN},1}$ and $x_{\text{TN},0}$, we have similar inequality regarding the upper-bound of robustness against accuracy attack:
$$
F(x_{\text{TN},0}) \le \min_{x_{\text{TN},1} \in \mathbb{S}_{01}} \xi_{\text{TN}, 1} +\sum_{t=1}^T \sqrt{n}\alpha Kd(x^{t-1,\text{Acc}}_{\text{TN},0},x^{t-1,\text{Acc}}_{\text{TN},1}).
$$
\label{rm: eod-acr2}
\end{remark}

\subsection{Fair adversarial training}

One direct result regarding Theorem \ref{acr-eod} is to incorporate adversarial samples w.r.t. accuracy attack during training to obtain a classifier that is also robust to adversarial fairness perturbations. Consider the relaxed DI loss under fairness perturbation: 
\begin{equation}
\begin{aligned}
&L'_{DI}\\
=&\left| \sum_{x_i \in \mathbb{S}_{.1}} \frac{f(x^{\text{DI}}_i)}{|\mathbb{S}_{.1}|} - \sum_{x_i \in \mathbb{S}_{.0}}\frac{f(x^{\text{DI}}_i)}{|\mathbb{S}_{.0}|}\right| \\
\leq &\left|\sum_{x_i\in \mathbb{S}_{.0}}\frac{f(x_i)}{|\mathbb{S}_{.0}|} - \sum_{x_i\in \mathbb{S}_{.1}}\frac{f(x_i)}{|\mathbb{S}_{.1}|}\right| + \sum_{a}\sum_{x_i\in \mathbb{S}_{.a}}\frac{\xi_i}{|\mathbb{S}_{.a}|}\\
= &L_{DI} + \sum_{a}\sum_{x_i\in \mathbb{S}_{.a}}\frac{\xi_i}{|\mathbb{S}_{.a}|},
\end{aligned}
\label{fair_adv}
\end{equation}
where $\xi_i$ is the change of $f(x_i)$ under $\epsilon$-level DI attack.
\eqref{fair_adv} shows that the relaxed DI loss under adversarial perturbation is upper-bounded by the relaxed DI loss without adversarial perturbation and robustness of samples against DI adversarial attack. {According to Theorem \ref{acr-eod} and \ref{rm: acr-eod2}, accuracy robustness upper bounds the robustness difference between samples against fairness attack. Since fairness and accuracy robustness are in alignment regarding samples in the `Alignment' subgroups,} a direct implication of this formulation is to improve robustness of classifier w.r.t. fairness by {incorporating accuracy adversarial samples} and fairness constraints during training:

\begin{equation}
\label{eq:obj}
\argmin_{\theta} \frac{1}{N} \sum_{i=1}^N L_\text{CE}(f(x^\text{Acc}_{i}),y_i),\; s.t.\;L \leq \gamma,
\end{equation}
where $x^\text{Acc}_{i}$ is the perturbed sample of $x_i$ after accuracy attack, and $L$ can be specified by fairness relaxations as regularization during training, or can be implicitly specified as preprocessing or postprocessing techniques.

Similarly, according to Theorem \ref{eod-acr} and \ref{rm: eod-acr2}, the change of predictions under accuracy attack is upper-bounded by the robustness against fairness attack, and it is feasible to improve robustness of classifier w.r.t. accuracy by using fairness adversarial samples during training.

\section{Experiments}

\begin{figure*}[!t]
    \centering
    \subfigure[EOd (Adult)]{\includegraphics[width=0.24\linewidth]{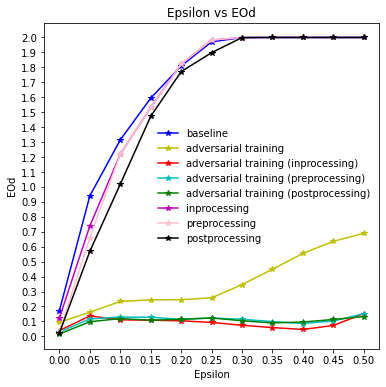}}
    \subfigure[DI (Adult)]{\includegraphics[width=0.24\linewidth]{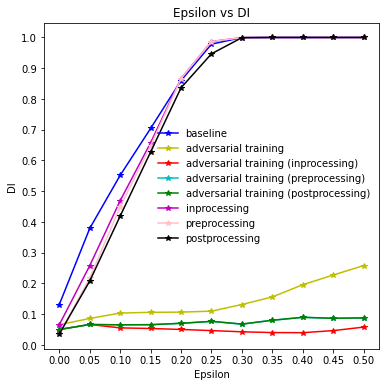}}
    \subfigure[Accuracy (Adult)]{\includegraphics[width=0.24\linewidth]{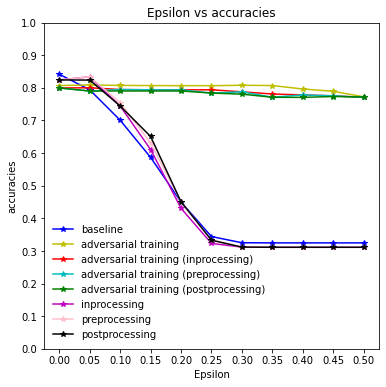}}
    \\
    \subfigure[EOd (COMPAS)]{\includegraphics[width=0.24\linewidth]{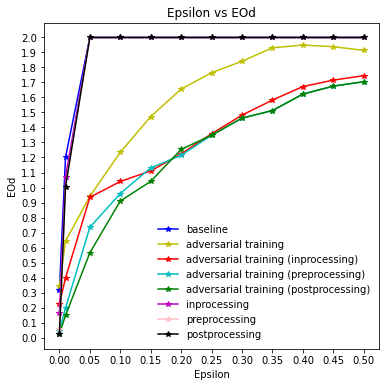}}
    \subfigure[DI (COMPAS)]{\includegraphics[width=0.24\linewidth]{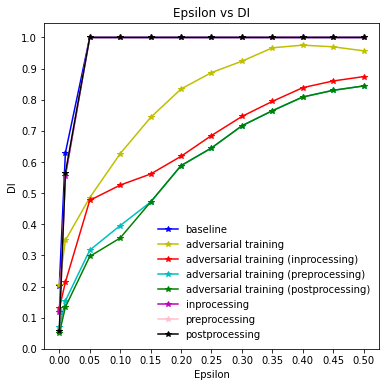}}
    \subfigure[Accuracy (COMPAS)]{\includegraphics[width=0.24\linewidth]{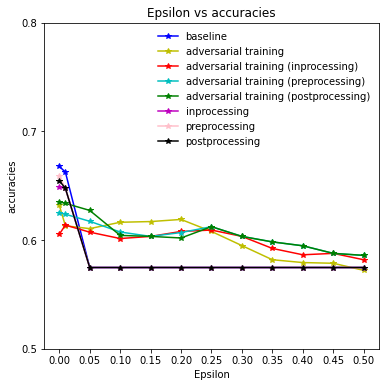}}
    \\
    \subfigure[EOd (German)]{\includegraphics[width=0.24\linewidth]{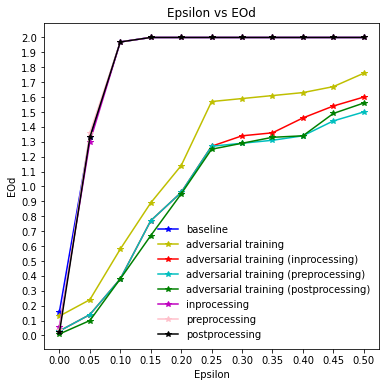}}
    \subfigure[DI (German)]{\includegraphics[width=0.24\linewidth]{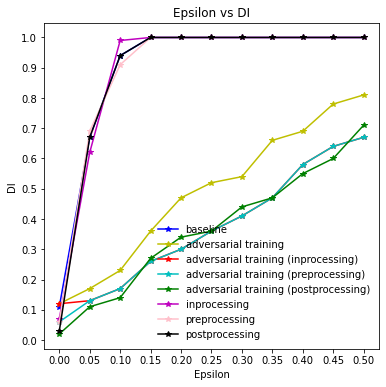}}
    \subfigure[Accuracy (German)]{\includegraphics[width=0.24\linewidth]{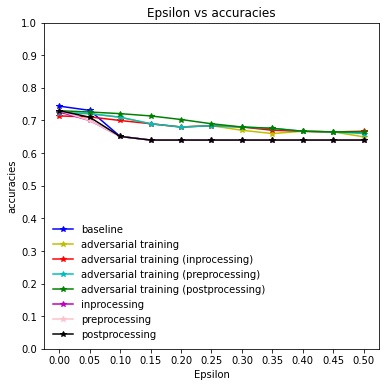}}
    \\
    \subfigure[EOd (CelebA)]{\includegraphics[width=0.24\linewidth]{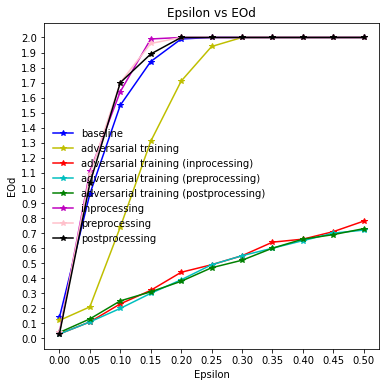}}
    \subfigure[DI (CelebA)]{\includegraphics[width=0.24\linewidth]{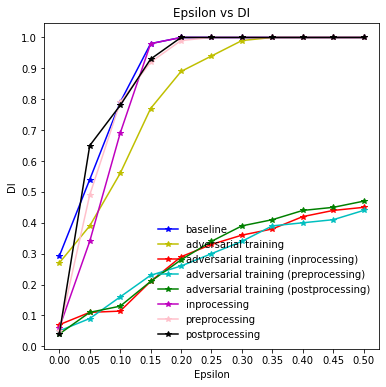}}
    \subfigure[Accuracy (CelebA)]{\includegraphics[width=0.24\linewidth]{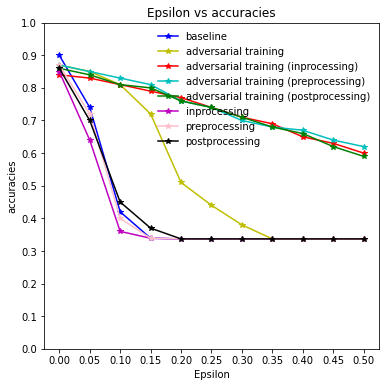}}
    \vspace{-10pt}
    \caption{Change in accuracy, DI and EOd under DI attack on four datasets. Our adversarial training methods (preprocessing, inprocessing, postprocessing) obtain improved fairness (lower EOd and DI) and higher accuracy with significant margin.}
    \label{change_disp}
\end{figure*}

\begin{figure*}[!t]
    \centering
    \subfigure[Black TPR (Adult)]{\includegraphics[width=0.24\linewidth]{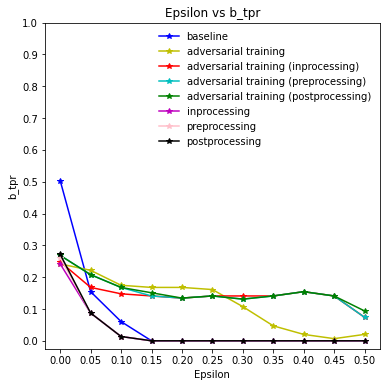}}
    \subfigure[White TPR (Adult)]{\includegraphics[width=0.24\linewidth]{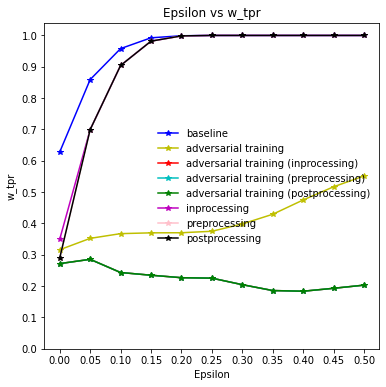}}
    \subfigure[Black TNR (Adult)]{\includegraphics[width=0.24\linewidth]{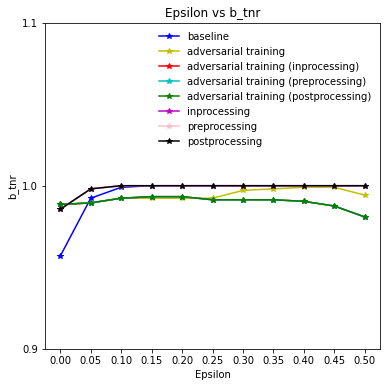}}
    \subfigure[White TNR (Adult)]{\includegraphics[width=0.24\linewidth]{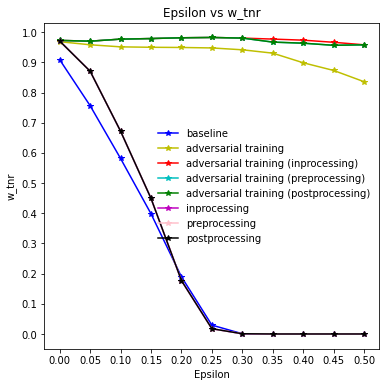}}
    \subfigure[Black TPR (COMPAS)]{\includegraphics[width=0.24\linewidth]{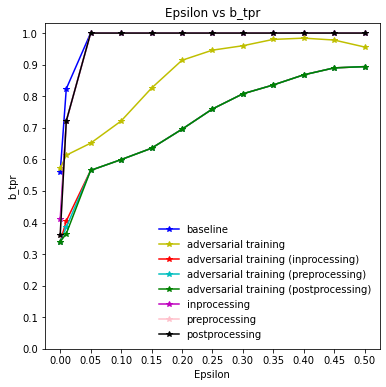}}
    \subfigure[White TPR (COMPAS)]{\includegraphics[width=0.24\linewidth]{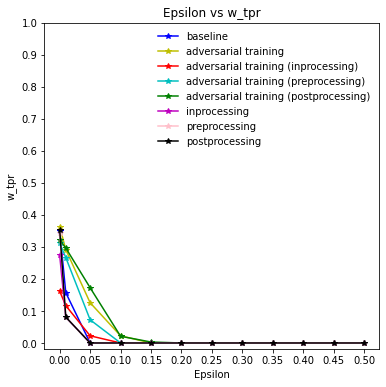}}
    \subfigure[Black TNR (COMPAS)]{\includegraphics[width=0.24\linewidth]{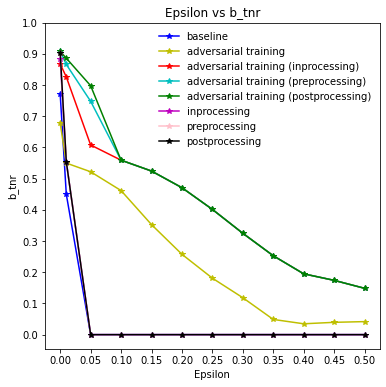}}
    \subfigure[White TNR (COMPAS)]{\includegraphics[width=0.24\linewidth]{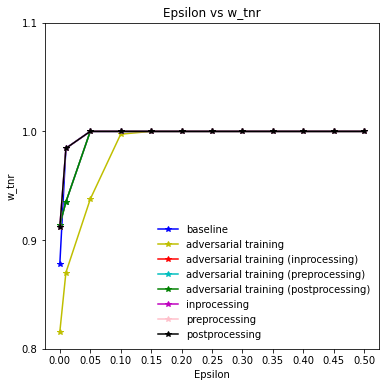}}
    \subfigure[Male TPR (German)]{\includegraphics[width=0.24\linewidth]{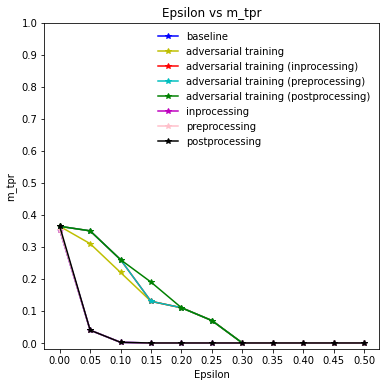}}
    \subfigure[Female TPR (German)]{\includegraphics[width=0.24\linewidth]{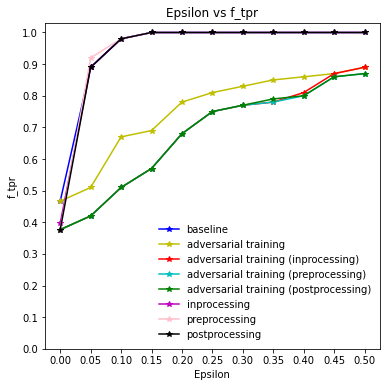}}
    \subfigure[Male TNR (German)]{\includegraphics[width=0.24\linewidth]{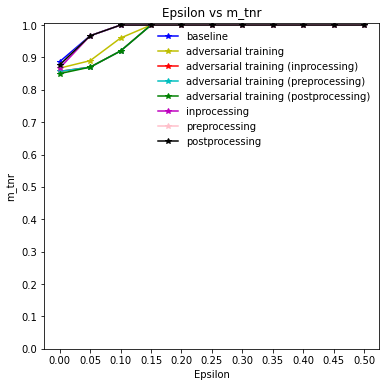}}
    \subfigure[Female TNR (German)]{\includegraphics[width=0.24\linewidth]{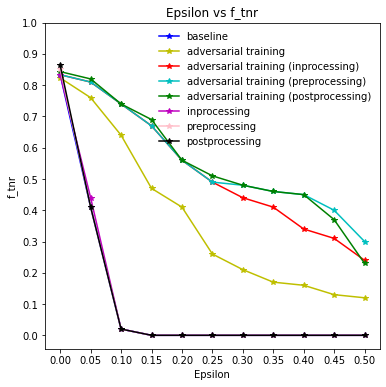}}
    \subfigure[Young TPR (CelebA)]{\includegraphics[width=0.24\linewidth]{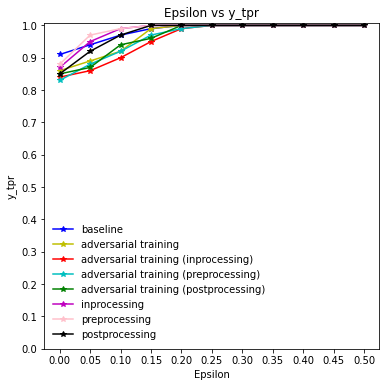}}
    \subfigure[Young TNR (CelebA)]{\includegraphics[width=0.24\linewidth]{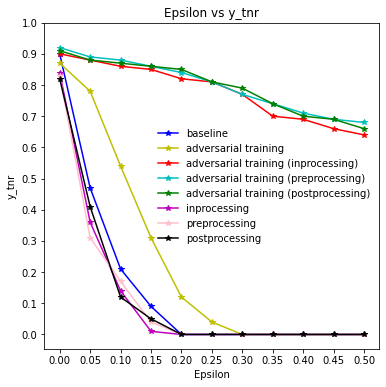}}
    \subfigure[Elder TPR (CelebA)]{\includegraphics[width=0.24\linewidth]{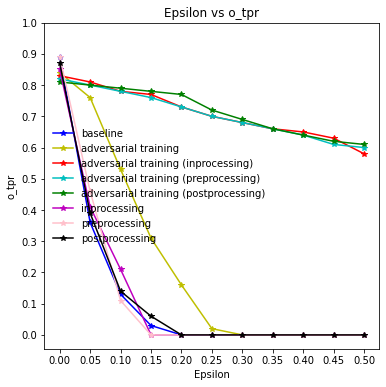}}
    \subfigure[Elder TNR (CelebA)]{\includegraphics[width=0.24\linewidth]{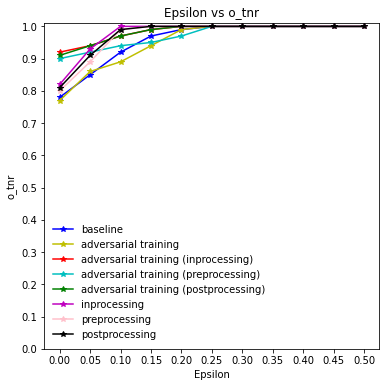}}
    \vspace{-10pt}
    \caption{Change of true positive rate (TPR) and true negative rate (TNR) under DI attack on four datasets.}
    \label{change_rate}
\end{figure*}

\begin{figure*}[!t]
    \centering
    \subfigure[EOd (adv)]{\includegraphics[width=0.24\linewidth]{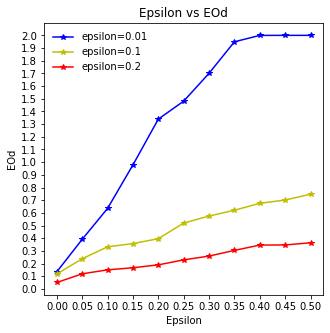}}
    \subfigure[EOd (adv+pre)]{\includegraphics[width=0.24\linewidth]{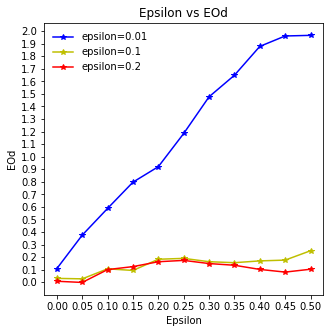}}
    \subfigure[EOd (adv+in)]{\includegraphics[width=0.24\linewidth]{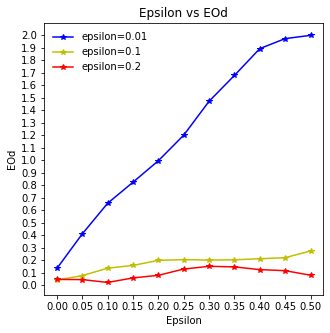}}
    \subfigure[EOd (adv+post)]{\includegraphics[width=0.24\linewidth]{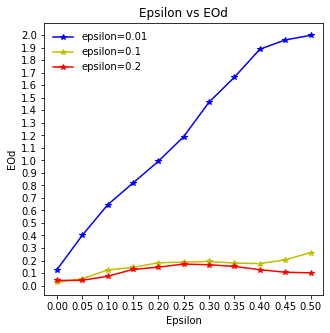}}
    \subfigure[DI (adv)]{\includegraphics[width=0.24\linewidth]{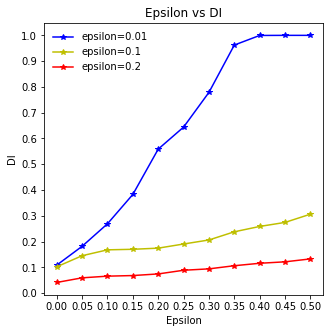}}
    \subfigure[DI (adv+pre)]{\includegraphics[width=0.24\linewidth]{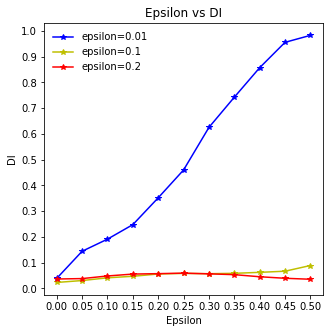}}
    \subfigure[DI (adv+in)]{\includegraphics[width=0.24\linewidth]{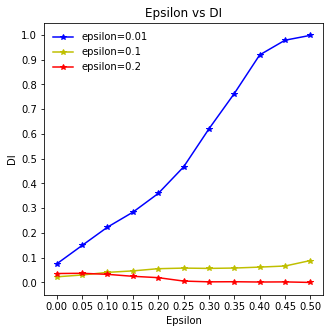}}
    \subfigure[DI (adv+post)]{\includegraphics[width=0.24\linewidth]{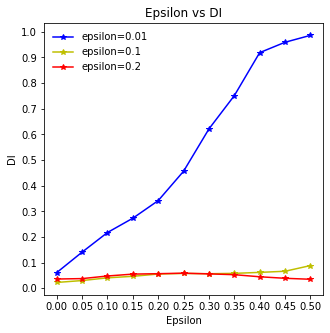}}
    \subfigure[Accuracy (adv)]{\includegraphics[width=0.24\linewidth]{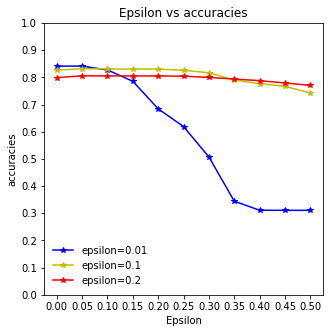}}
    \subfigure[Accuracy (adv+pre)]{\includegraphics[width=0.24\linewidth]{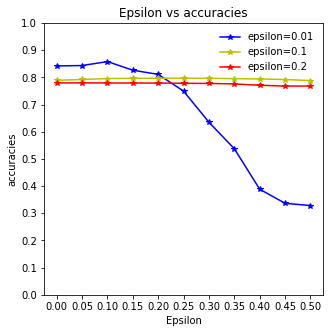}}
    \subfigure[Accuracy (adv+in)]{\includegraphics[width=0.24\linewidth]{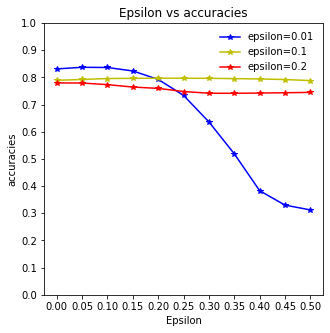}}
    \subfigure[Accuracy (adv+post)]{\includegraphics[width=0.24\linewidth]{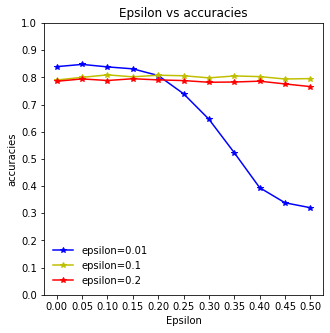}}
    \vspace{-10pt}
    \caption{Change of accuracy, DI and EOd under DI attack with varying training perturbation $\epsilon$ on Adult dataset.}
    \vspace{-10pt}
    \label{ep_adv_adt}
\end{figure*}

\begin{figure*}[!t]
    \centering
    \subfigure[Accuracy]{\includegraphics[width=0.24\linewidth]{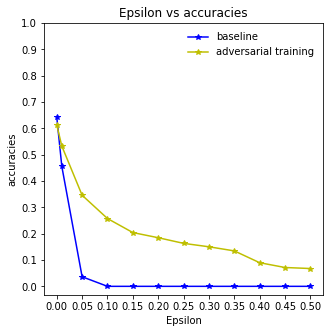}}
    \subfigure[DI]{\includegraphics[width=0.24\linewidth]{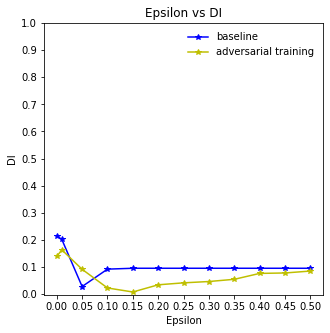}}
    \subfigure[EOd]{\includegraphics[width=0.24\linewidth]{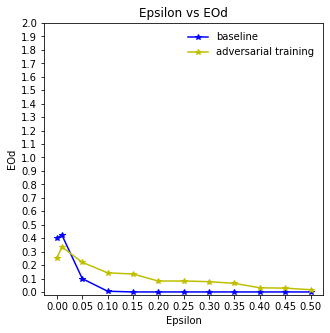}}
    \vspace{-10pt}
\caption{Results of a classifier adversarially trained w.r.t. DI. Change of accuracy, DI and EOd under accuracy attack on Adult dataset.}
\vspace{-10pt}
    \label{acr_adv_adt}
\end{figure*}

We evaluate our method on four datasets: 

\begin{itemize}
\item Adult \citep{Dua:2019}: The Adult dataset contains 65,123 samples with 14 attributes. The goal is to predict whether an individual’s annual income exceeds $50K$, and the sensitive attribute is chosen as \emph{race}.

\item COMPAS \citep{larson2016compas}: The ProPublica COMPAS dataset contains 7,215 samples with 10 attributes. The goal is to predict whether a defendant re-offend within two years. Following the protocol in earlier fairness methods \citep{zafar2017fairness}, we only select white and black individuals in COMPAS dataset, which contains 6,150 samples in total. The sensitive attribute in this dataset is \emph{race}.

\item German \citep{Dua:2019}: The German credit risk dateset  contains 1,000 samples with 9 attributes. The goal is to predict whether a client is highly risky, and the sensitive attribute in this dataset is \emph{sex}.

\item CelebA \citep{liu2015faceattributes}: CelebA dataset contains 202,599 samples with 40 binary attributes. We choose gender as target label, and the sensitive attribute in this dataset is \emph{age}. 
\end{itemize}

We use accuracy as performance evaluation, and disparate impact (DI) and equalized odds (EOd) as fairness metric. The classifier for all compared methods is chosen as MLP, and all methods are trained under the same data partition. During adversarial training, the perturbation level is set as 0.2 for Adult dataset, 0.005 for COMPAS dataset, 0.01 for German dataset and 0.1 for CelebA dataset, where the the perturbation level is empirically determined to achieve the largest perturbation while still ensuring convergence.
\subsection{Robustness against fairness attack}
We compare five different methods with our fair adversarial training method. Specifically, we consider three different versions for our fair adversarial training method (preprocessing, inprocessing and postprocessing). The three versions differ in the fairness regularization $L$ in \eqref{eq:obj}.
\begin{itemize}
    \item Baseline: MLP model under normal training.
    \item Preprocessing: MLP model under normal training with label processed by \citet{jiang2020identifying}.
   \item
   Inprocessing: MLP model under normal training with relaxed EOd constraint by \citet{wang2022understanding}.
   \item
   Postprocessing: MLP model under normal training with postprocessing technique by \citet{jang2022group}.
    \item Adversarial training: MLP model under adversarial training w.r.t. accuracy.
    \item Adversarial training (preprocessing): MLP model {under adversarial training w.r.t. accuracy} with training label processed by \citet{jiang2020identifying}.
    \item Adversarial training (inprocessing): MLP model under adversarial training w.r.t. accuracy with relaxed EOd constraint by \citet{wang2022understanding}.
    \item Adversarial training (postprocessing): MLP model under adversarial training w.r.t. accuracy with predicted label postprocessed by \citet{jang2022group}.
\end{itemize}

Results on classifiers under DI attack are shown in Fig. \ref{change_disp} - \ref{change_rate}. The DI attack enforces biased predictions against testing samples based on the sensitive information, and under a successful attack (the DI reaches its maximum), the EOd also reaches its maximum, while the accuracy under adversarial attack is determined by the base rate of sensitive groups. Compared with methods under adversarial training, methods under normal training show a sharp increase in DI and EOd under adversarial attack, and improvement in fairness of classifiers under normal training do not help with the robustness under adversarial perturbation. In comparison, classifiers under adversarial training w.r.t. accuracy show improvement in terms of robustness against fairness attack, and classifiers under fair adversarial training show further remarkable improvement in terms of robustness against fairness attack\footnote{We defer the detailed values for fair adversarial training in Appendix, as part of the fair adversarial training results are overlapped with each other.}. Besides, for classifiers under adversarial training w.r.t. accuracy, the TPR and TNR of different groups show slower change compared with those of classifiers under normal training. {This shows that improving robustness against accuracy attack also improves robustness against fairness attack and is in line with our discussion in \eqref{fair_adv}.}

Furthermore, Fig. \ref{ep_adv_adt} shows the effect of perturbation levels during training on Adult dataset, where the results show that larger perturbation level during training indicates better robustness against fairness adversarial attack for both vanilla adversarial training and fair adversarial training during testing. We defer full results of varying training perturbation levels to appendix.

\subsection{Robustness against accuracy attack}
We move on to discuss the improvement of robustness w.r.t. accuracy under adversarial training w.r.t. fairness. We compare two different methods:
\begin{itemize}
    \item Baseline: MLP model under normal training.
    \item Adversarial training (DI): MLP model under adversarial training w.r.t. relaxed DI.
\end{itemize}
We show results on classifiers under accuracy attack on Adult dataset in Fig. \ref{acr_adv_adt}.
Under a successful accuracy attack (the accuracy reaches its minimum), the EOd also becomes zero, while DI does not necessarily vanishes due to distributional disparities across sensitive groups. Compared with baseline classifier, classifier under adversarial training w.r.t. DI shows remarkable improvement in robustness against accuracy attack. This shows that robustness against accuracy attack also benefits from adversarial training against fairness. Results on other datasets are shown in the appendix.

\section{Conclusion}
Fairness attack and fairness adversarial training is an important yet not properly addressed problem. In this paper, we propose a unified framework for fairness attack against group fairness notions, where we show theoretically the alignment of attack against different notions, and we demonstrate the connections between fairness attack and conventional accuracy attack. We show theoretically the alignment between accuracy robustness and fairness robustness, and we propose a fair adversarial training structure, where the goal is to improve adversarial robustness w.r.t. accuracy while ensuring fairness. Further, from experiments we validate that our method achieves better robustness under fairness adversarial attack, and that the robustness w.r.t. fairness and accuracy align with each other. Future directions include finding alternative relaxations for fairness attack, and alternative training strategies for fair adversarial training. 

\bibliography{ref}
\bibliographystyle{hapalike}

\newpage
\appendix
\onecolumn
\section{Appendix}

\subsection{Full results on varying $\epsilon$}

\begin{figure*}[!tbph]
    \centering
    \subfigure[EOd (adv)]{\includegraphics[width=0.24\linewidth]{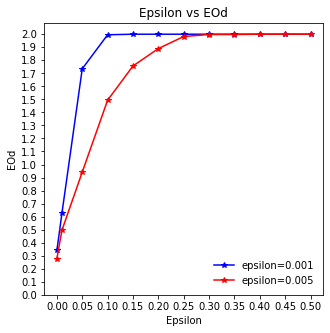}}
    \subfigure[EOd (adv+pre)]{\includegraphics[width=0.24\linewidth]{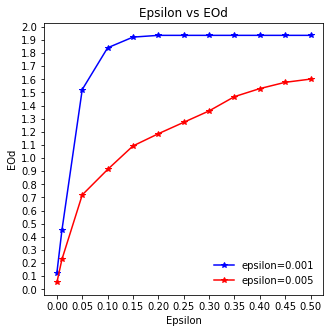}}
    \subfigure[EOd (adv+in)]{\includegraphics[width=0.24\linewidth]{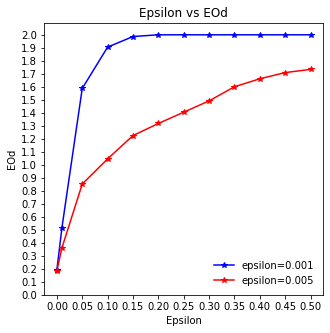}}
    \subfigure[EOd (adv+post)]{\includegraphics[width=0.24\linewidth]{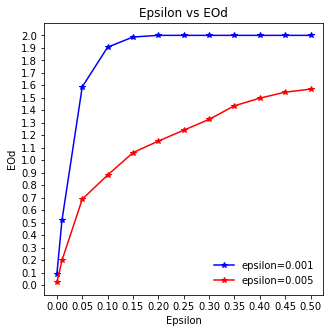}}
    \subfigure[DI (adv)]{\includegraphics[width=0.24\linewidth]{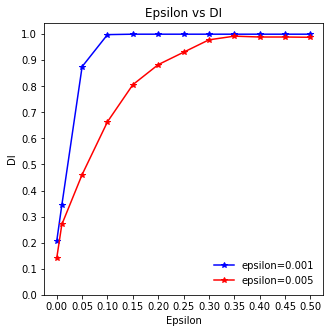}}
    \subfigure[DI (adv+pre)]{\includegraphics[width=0.24\linewidth]{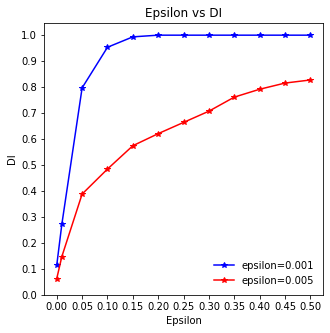}}
    \subfigure[DI (adv+in)]{\includegraphics[width=0.24\linewidth]{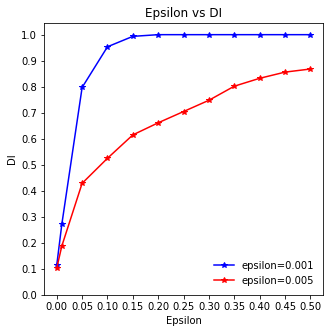}}
    \subfigure[DI (adv+post)]{\includegraphics[width=0.24\linewidth]{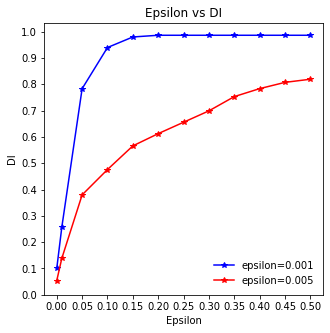}}
    \subfigure[Acr (adv)]{\includegraphics[width=0.24\linewidth]{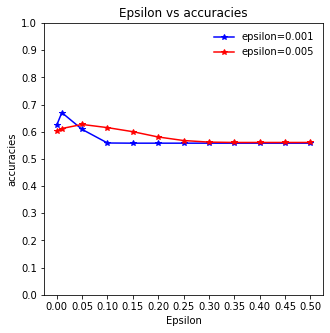}}
    \subfigure[Acr (adv+pre)]{\includegraphics[width=0.24\linewidth]{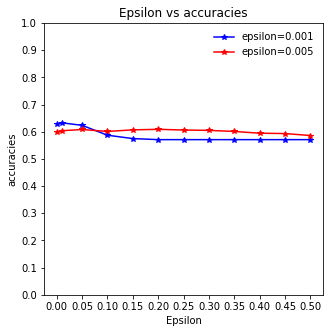}}
    \subfigure[Acr (adv+in)]{\includegraphics[width=0.24\linewidth]{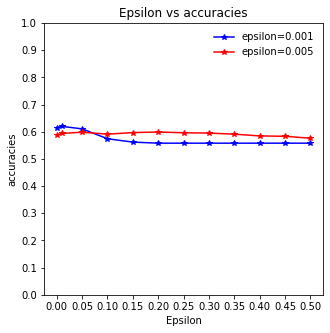}}
    \subfigure[Acr (adv+post)]{\includegraphics[width=0.24\linewidth]{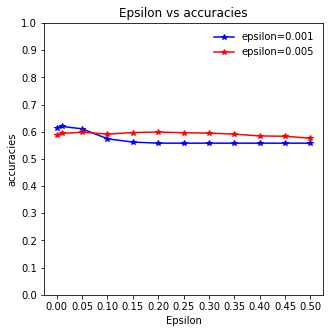}}
    \caption{Change of accuracy, DI and EOd under DI attack with varying training perturbation $\epsilon$ on COMPAS dataset.}
    \label{ep_adv_cps}
\end{figure*}

\begin{figure*}[!h]
    \centering
    \subfigure[EOd (adv)]{\includegraphics[width=0.23\linewidth]{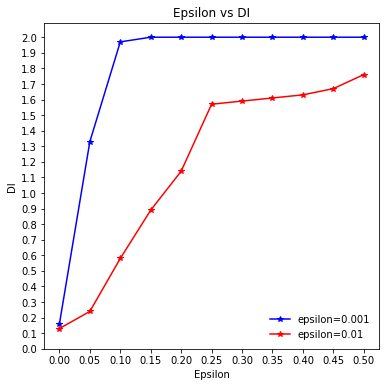}}
    \subfigure[EOd (adv+pre)]{\includegraphics[width=0.23\linewidth]{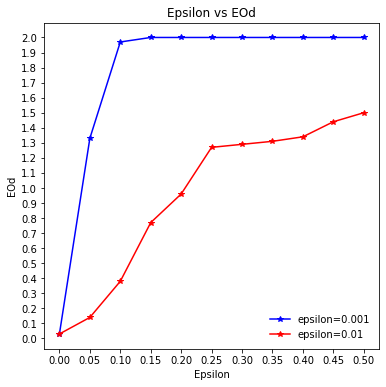}}
    \subfigure[EOd (adv+in)]{\includegraphics[width=0.23\linewidth]{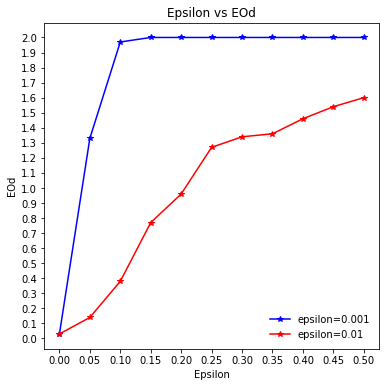}}
    \subfigure[EOd (adv+post)]{\includegraphics[width=0.23\linewidth]{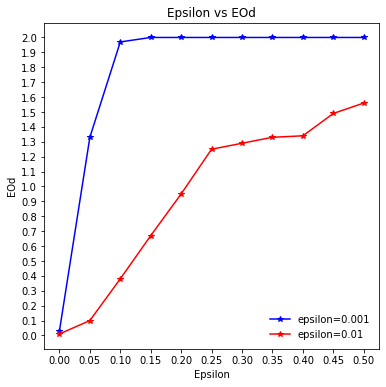}}
    \subfigure[DI (adv)]{\includegraphics[width=0.23\linewidth]{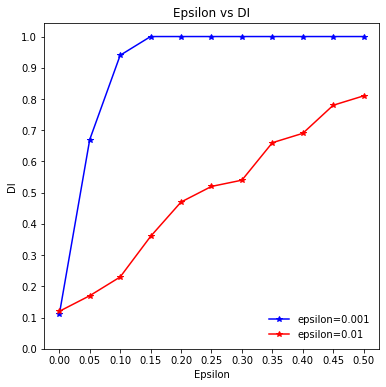}}
    \subfigure[DI (adv+pre)]{\includegraphics[width=0.23\linewidth]{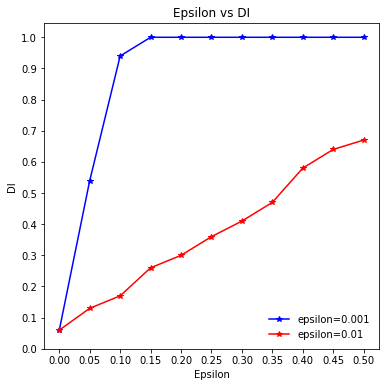}}
    \subfigure[DI (adv+in)]{\includegraphics[width=0.23\linewidth]{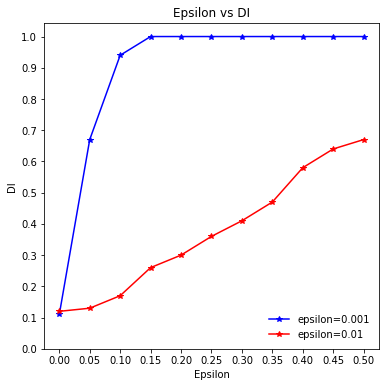}}
    \subfigure[DI (adv+post)]{\includegraphics[width=0.23\linewidth]{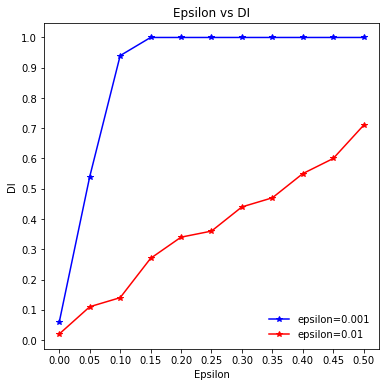}}
    \subfigure[Acr (adv)]{\includegraphics[width=0.23\linewidth]{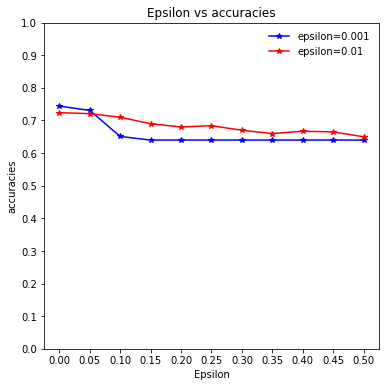}}
    \subfigure[Acr (adv+pre)]{\includegraphics[width=0.23\linewidth]{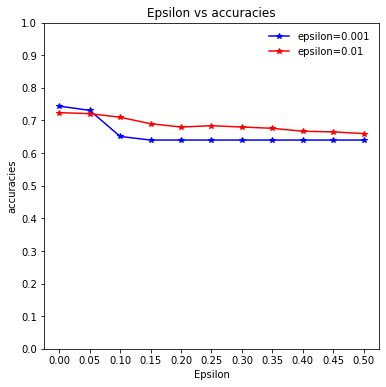}}
    \subfigure[Acr (adv+in)]{\includegraphics[width=0.23\linewidth]{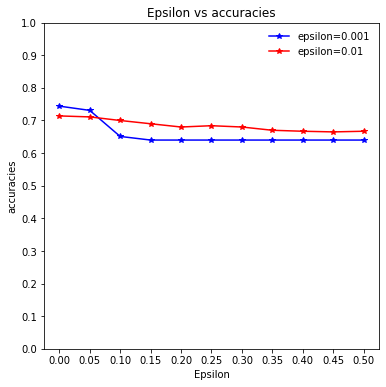}}
    \subfigure[Acr (adv+post)]{\includegraphics[width=0.23\linewidth]{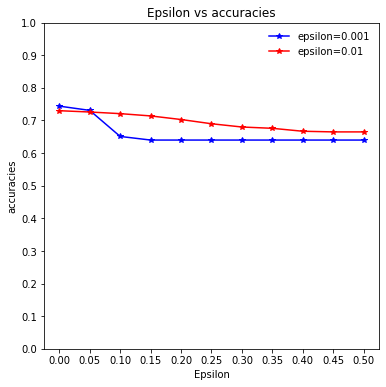}}
    \caption{Change of accuracy, DI and EOd under DI attack with varying training perturbation $\epsilon$ on German dataset.}
    \label{ep_adv_german}
\end{figure*}

\begin{figure*}[!h]
    \centering
    \subfigure[EOd (adv)]{\includegraphics[width=0.23\linewidth]{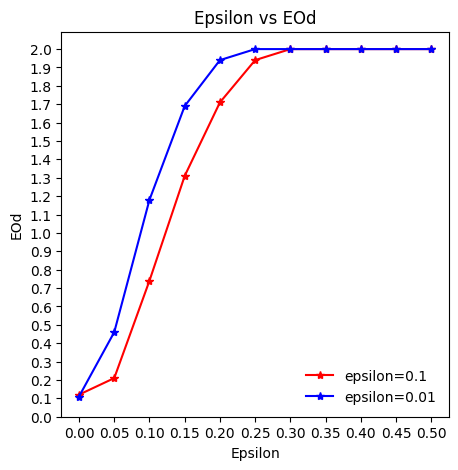}}
    \subfigure[EOd (adv+pre)]{\includegraphics[width=0.23\linewidth]{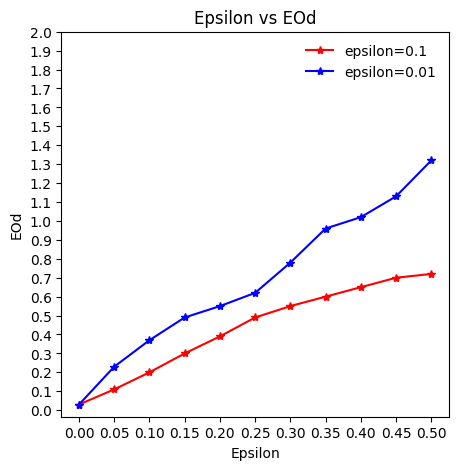}}
    \subfigure[EOd (adv+in)]{\includegraphics[width=0.23\linewidth]{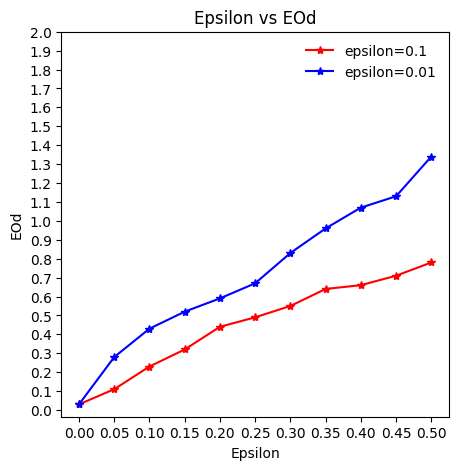}}
    \subfigure[EOd (adv+post)]{\includegraphics[width=0.23\linewidth]{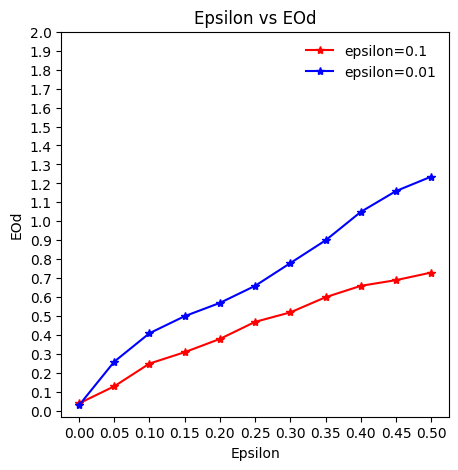}}
    \subfigure[DI (adv)]{\includegraphics[width=0.23\linewidth]{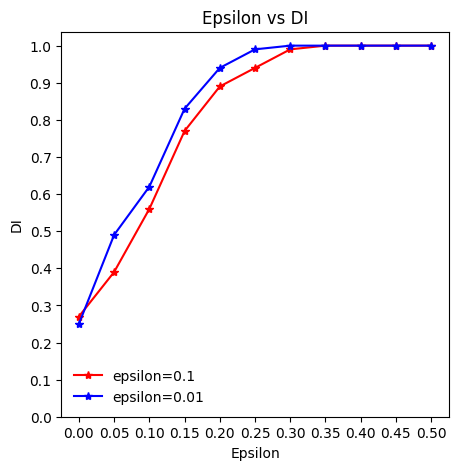}}
    \subfigure[DI (adv+pre)]{\includegraphics[width=0.23\linewidth]{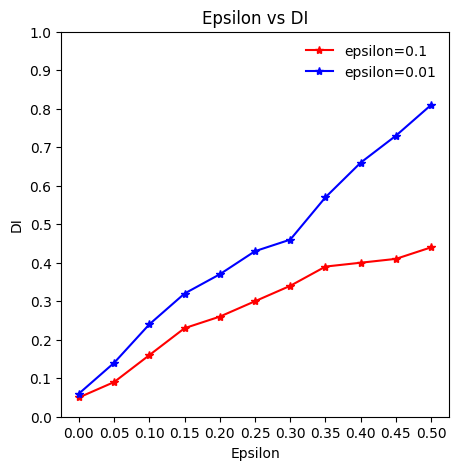}}
    \subfigure[DI (adv+in)]{\includegraphics[width=0.23\linewidth]{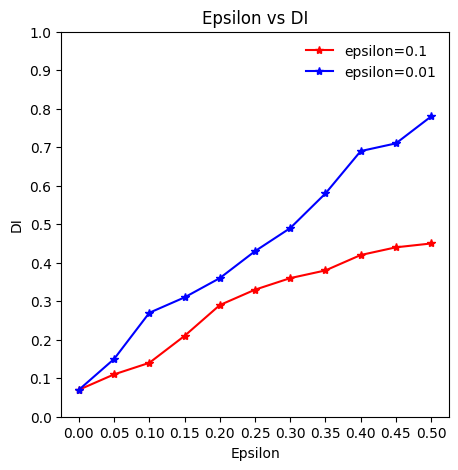}}
    \subfigure[DI (adv+post)]{\includegraphics[width=0.23\linewidth]{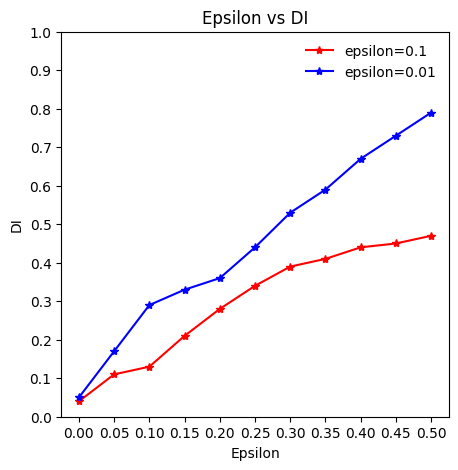}}
    \subfigure[Acr (adv)]{\includegraphics[width=0.23\linewidth]{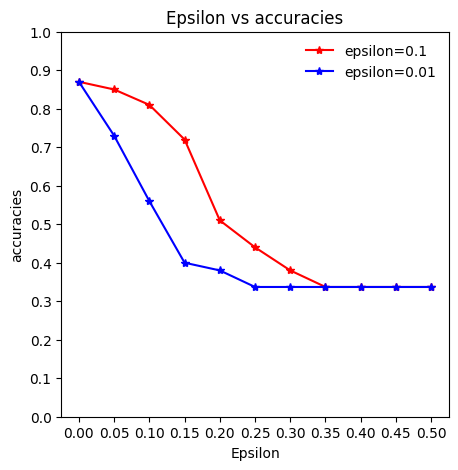}}
    \subfigure[Acr (adv+pre)]{\includegraphics[width=0.23\linewidth]{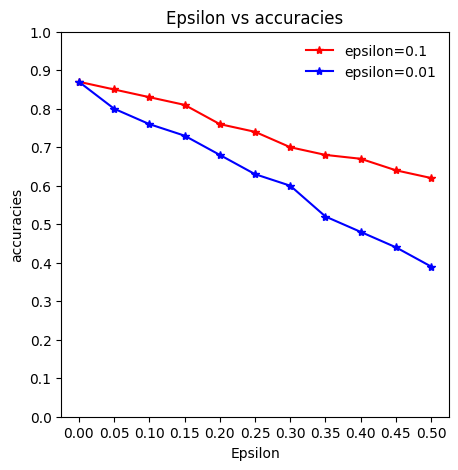}}
    \subfigure[Acr (adv+in)]{\includegraphics[width=0.23\linewidth]{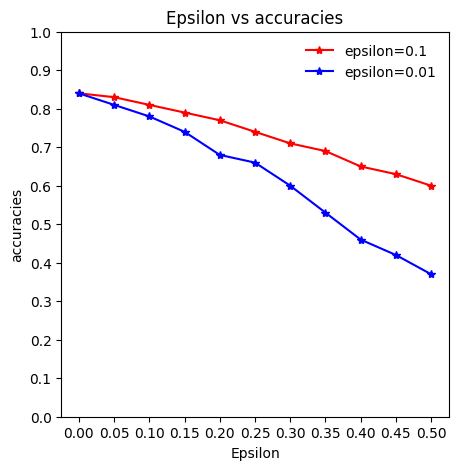}}
    \subfigure[Acr (adv+post)]{\includegraphics[width=0.23\linewidth]{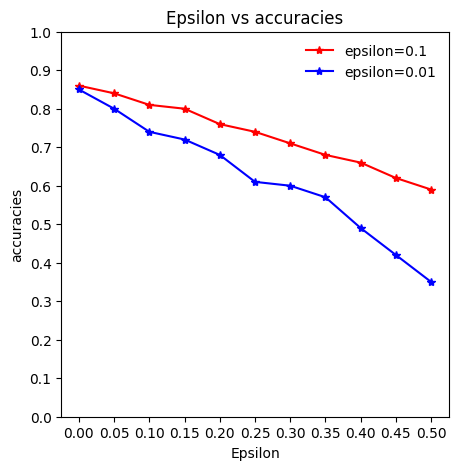}}
    \caption{Change of accuracy, DI and EOd under DI attack with varying training perturbation $\epsilon$ on CelebA dataset.}
    \label{ep_adv_celeba}
\end{figure*}

Results of varying $\epsilon$ on COMPAS, German and CelebA dataset can be found in Fig. \ref{ep_adv_cps}-\ref{ep_adv_celeba}. As shown in the figures, larger perturbation levels result in classifiers that are more robust to adversarial perturbations.

\subsection{Proof of Corollary \ref{cor_1}}

\begin{proof}
The objective for EOd attack can be written as the following form:
\begin{equation*}
\begin{aligned}
L_{EOd} &= \left|\sum_{x_i \in \mathbb{S}_{00}}\frac{f(x_i)}{|\mathbb{S}_{00}|}-\sum_{x_i \in \mathbb{S}_{01}} \frac{f(x_i)}{|\mathbb{S}_{01}|}\right| + \left|\sum_{x_i \in \mathbb{S}_{10}}\frac{f(x_i)}{|\mathbb{S}_{10}|}-\sum_{x_i \in \mathbb{S}_{11}} \frac{f(x_i)}{|\mathbb{S}_{11}|}\right|
\\
& \ge \left|\sum_{x \in \mathbb{S}_{00}}\frac{f(x)}{|\mathbb{S}_{00}|} - \sum_{x \in \mathbb{S}_{01}}\frac{f(x)}{|\mathbb{S}_{01}|} + \sum_{x \in \mathbb{S}_{10}}\frac{f(x)}{|\mathbb{S}_{10}|} - \sum_{x \in \mathbb{S}_{11}}\frac{f(x)}{|\mathbb{S}_{11}|}\right|
\\
& {=} \left|\sum_{x \in \mathbb{S}_{00}}\frac{|\mathbb{S}_{.0}|}{|\mathbb{S}_{00}|}\frac{f(x)}{|\mathbb{S}_{.0}|} + \sum_{x \in \mathbb{S}_{10}}\frac{|\mathbb{S}_{.0}|}{|\mathbb{S}_{10}|}\frac{f(x)}{|\mathbb{S}_{.0}|} - \sum_{x \in \mathbb{S}_{01}}\frac{|\mathbb{S}_{.1}|}{|\mathbb{S}_{01}|}\frac{f(x)}{|\mathbb{S}_{.1}|} - \sum_{x \in \mathbb{S}_{11}}\frac{|\mathbb{S}_{.1}|}{|\mathbb{S}_{11}|}\frac{f(x)}{|\mathbb{S}_{.1}|}\right|.
\end{aligned}
\end{equation*}

This shows that the EOd attack is lower-bounded by the weighted DI attack as in \eqref{eq:DP}. 
Specifically, under a successful DI attack, we have $f(x)=1, \forall x \in \mathbb{S}_{.a}$ and $f(x)=0, \forall x \in \mathbb{S}_{.a'}$, and the lower bound can be simplified as 
\begin{equation}
L_{\text{EOd}} \ge { {\left|\sum_{x \in \mathbb{S}_{0a}}\frac{|\mathbb{S}_{.a}|}{|\mathbb{S}_{0a}|}\frac{1}{|\mathbb{S}_{.a}|} + \sum_{x \in \mathbb{S}_{1a}}\frac{|\mathbb{S}_{.a}|}{|\mathbb{S}_{1a}|}\frac{1}{|\mathbb{S}_{.a}|}\right|}   } = 2,
\end{equation}
which shows that a successful DI attack always implies a successful EOd attack.

\begin{remark}
A successful EOd attack does not always imply a successful DI attack. Assume $\sum_{x_i \in \mathbb{S}_{00}}\frac{f(x_i)}{|\mathbb{S}_{00}|} \le \sum_{x_i \in \mathbb{S}_{01}} \frac{f(x_i)}{|\mathbb{S}_{01}|}$ and $\sum_{x_i \in \mathbb{S}_{10}}\frac{f(x_i)}{|\mathbb{S}_{10}|} \ge \sum_{x_i \in \mathbb{S}_{11}} \frac{f(x_i)}{|\mathbb{S}_{11}|}$, under a successful EOd attack, all the predictions in the disadvantaged group will become correct, while all the predictions in the advantaged group will become incorrect, and the disparate impact will not be maximized as both groups contain positive predictions.
\end{remark}
\end{proof}

\subsection{Proof of Corollary \ref{cor_2}}
\begin{proof}
The objective for accuracy attack for sample $x_i$ can be written as 
\begin{equation}
\max_{\delta} L((x_i+\epsilon),y_i), \|\epsilon\| \le \epsilon_0,
\label{acr_atk}
\end{equation}
Consider the DI attack in \eqref{eq:DP}, we have the objective for DI attack as follows:
\begin{equation*}
\max_{\delta} \alpha_{i}  \frac{f(x_i+\epsilon)}{|\mathbb{S}_{.a_i}|}, \|\epsilon\| \le \epsilon',
\end{equation*}
where $\alpha_{i} = -1$ for $a_i=0$ and $\alpha_{i} = 1$ for $a_i=1$. For positive samples, we can further write \eqref{acr_atk} as
\begin{equation*}
\max_{\delta} -\log(f(x_i+\epsilon)), \|\epsilon\| \le \epsilon_0,
\end{equation*}
where the perturbation is expected to minimize the predicted soft label, which is in alignment with the objective for DI when $\alpha_{i} = -1$, i.e., for TP and FN disadvantaged samples, the two attack are in alignment. Similarly, for negative samples, we have \eqref{acr_atk} as
\begin{equation*}
\max_{\delta} -\log(1-f(x+\epsilon)), \|\epsilon\| \le \epsilon',
\end{equation*}
where the perturbation is expected to maximize the predicted soft label, which is in alignment with the objective for DI when $\alpha_{i} = 1$, i.e., for TN and FP advantaged samples, the two attack are in alignment. Specifically, for gradient-based attack, we have the two kinds of attack equivalent.
\end{proof}
\subsection{Proof of Theorem \ref{acr-eod}}

\begin{proof}
Let $f$ be the function of classifier, consider the positive testing set $\{(x_i, 1, a_i), 1 \le i \le N\}$ for simplicity, at $t-1$-th iteration, we have the linear approximation of testing CE loss under DI attack as follows:
\begin{equation}
\label{eq:changeL}
L_\text{CE}(x^t) =-\log(x^t) = -\log(f(x^{t-1})-\delta^{t-1}) = -\log(f(x^{t-1})) + \frac{\delta^{t-1}}{f(x^{t-1})} + r_L(x^{t-1}),
\end{equation}
where $\delta^{t-1}$ is the change of soft label induced by DI attack at $t-1$-th iteration, and $r_L(x)$ is the remainder of Taylor's expansion. For gradient-based attack, the predicted soft label for adversarial sample can be formulated as 
\begin{equation}
\label{eq:changef}
\begin{aligned}
&f(x^t)
\\
=&f(x^{t-1} + \alpha\text{sign}(\nabla_{x^{t-1}}L_{DI}))
\\
=&f(x^{t-1}) + \alpha  (\nabla_{x^{t-1}}f(x^{t-1}))^T\text{sign}(\nabla_{x^{t-1}}L_{DI}) + r_f(x^{t-1}),
\end{aligned}
\end{equation}
where $\epsilon$ is the magnitude of perturbation, $L_{DI}$ is the relaxed DI loss, and $r_f(x)$ is the remainder of Taylor's expansion. Let $D(x^t) \coloneqq |L(x^{t})-L(x^{t-1})|$ be the change of CE loss under DI attack at $t$-th iteration, according to \eqref{eq:changeL} and \eqref{eq:changef} we have 
\begin{equation*}
\begin{aligned}
D(x^t) &= |L_\text{CE}(x^{t})-L_\text{CE}(x^{t-1})| \\
& = |-\log(f(x^{t-1})) + \frac{\delta^{t-1}}{f(x^{t-1})} + r_L(x) + \log(f(x))|
\\
&\approx \frac{|\alpha  (\nabla_{x^{t-1}}f(x^{t-1}))^T\text{sign}(\nabla_{x^{t-1}}L_{DI})|}{f(x^{t-1})}.
\end{aligned}
\end{equation*}
Consider FN sample $x_{\text{FN}, 0}$ from disadvantaged group and FN sample $x_{\text{FN},1}$ from advantaged group, since the gradient of $f$ w.r.t. $x$ is Lipschitz with constant $K$, we have the difference of change in  CE loss under DI attack at $t$-th iteration as follows:
\begin{equation}
\begin{aligned}
&|D(x_{\text{FN}, 1}^{t-1,\text{DI}})-D(x_{\text{FN}, 0}^{t-1,\text{DI}})|
\\
=&  \alpha  \left|\frac{|(\nabla_{x_{\text{FN}, 1}^{t-1,\text{DI}}}f(x_{\text{FN}, 1}^{t-1,\text{DI}}))^T	 \text{sign}(\nabla_{x_{\text{FN}, 1}^{t-1,\text{DI}}}L_f)|}{f(x_{\text{FN}, 1}^{t-1,\text{DI}})} -\frac{|(\nabla_{x_{\text{FN}, 0}^{t-1,\text{DI}}}f(x_{\text{FN}, 0}^{t-1,\text{DI}}))^T	 \text{sign}(\nabla_{x_{\text{FN}, 0}^{t-1,\text{DI}}}L_f)|}{f(x_{\text{FN}, 0}^{t-1,\text{DI}})}\right|
\\
=&  \alpha  \left|\frac{(\nabla_{x_{\text{FN}, 1}^{t-1,\text{DI}}}f(x_{\text{FN}, 1}^{t-1,\text{DI}}))^T	 \text{sign}(\nabla_{x_{\text{FN}, 1}^{t-1,\text{DI}}}L_f)}{f(x_{\text{FN}, 1}^{t-1,\text{DI}})} + \frac{(\nabla_{x_{\text{FN}, 0}^{t-1,\text{DI}}}f_{\theta}(x_{\text{FN}, 0}^{t-1,\text{DI}}))^T	 \text{sign}(\nabla_{x_{\text{FN}, 0}^{t-1,\text{DI}}}L_f)}{f(x_{\text{FN}, 0}^{t-1,\text{DI}})}\right|
\\
= & \alpha  \left|\frac{(\nabla_{x_{\text{FN}, 1}^{t-1,\text{DI}}}f(x_{\text{FN}, 1}^{t-1,\text{DI}}))^T\text{sign}(\frac{1}{N_1}\nabla_{x_{\text{FN}, 1}^{t-1,\text{DI}}}f(x_{\text{FN}, 1}^{t-1,\text{DI}}))}{f_{\theta}(x_{\text{FN}, 1}^{t-1,\text{DI}})}  - \frac{(\nabla_{x_{\text{FN}, 0}^{t-1,\text{DI}}}f(x_{\text{FN}, 0}^{t-1,\text{DI}}))^T\text{sign}(\frac{1}{N_0}\nabla_{x_{\text{FN}, 0}^{t-1,\text{DI}}}f(x_{\text{FN}, 0}^{t-1,\text{DI}}))}{f(x_{\text{FN}, 0}^{t-1,\text{DI}})}\right|
\\
= & \alpha  \left|\frac{\sum_{j=1}^n |\partial_{x_j} f(x_{\text{FN}, 1}^{t-1,\text{DI}})|}{f(x_{\text{FN}, 1}^{t-1,\text{DI}})}  - \frac{\sum_{j=1}^n |\partial_{x_j} f(x_{\text{FN}, 0}^{t-1,\text{DI}})|}{f(x_{\text{FN}, 0}^{t-1,\text{DI}})}\right|
\\
= & \alpha  \left|\frac{\|\nabla_{x_{\text{FN}, 1}^{t-1,\text{DI}}}f(x_{\text{FN}, 1}^{t-1,\text{DI}})\|_1}{f(x_{\text{FN}, 1}^{t-1,\text{DI}})}  - \frac{\|\nabla_{x_{\text{FN}, 0}^{t-1,\text{DI}}}f(x_{\text{FN}, 0}^{t-1,\text{DI}})\|_1}{f(x_{\text{FN}, 0}^{t-1,\text{DI}})} \right|,
\end{aligned}
\label{eq:acc-eod}
\end{equation}

where $n$ is the dimension of input feature. Since $\nabla_{x}f(x)$ is Lipschitz, we have
\begin{equation*}
\|\nabla_{x}f(x_1)\|_2 - \|\nabla_{x}f(x_0)\|_2 \le \|\nabla_{x}f(x_1) - \nabla_{x}f(x_0)\|_2 \le Kd(x_1,x_0),
\end{equation*}
where the first sign is due to triangle inequality. By Jensen's inequality we  have $\|x\|_2 \le \|x\|_1 \leq \sqrt{n} \|x\|_2$, and
\begin{equation}
\|\nabla_{x}f(x_1)\|_1 - \|\nabla_{x}f(x_0)\|_1 \le \|\nabla_{x} f(x_0) - \nabla_{x} f(x_1)\|_1 \le \sqrt{n}Kd(x_1,x_0).
\label{ineq}
\end{equation}
Assume $\frac{\|\nabla_{x_{\text{FN}, 1}^{t-1,\text{DI}}}f(x_{\text{FN}, 1}^{t-1,\text{DI}})\|_1}{f(x_{\text{FN}, 1}^{t-1,\text{DI}})} \ge \frac{\|\nabla_{x_{\text{FN}, 0}^{t-1,\text{DI}}}f(x_{\text{FN}, 0}^{t-1,\text{DI}})\|_1}{f(x_{\text{FN}, 0}^{t-1,\text{DI}})}$, plugging \eqref{ineq} back into \eqref{eq:acc-eod}, we have 
\begin{equation}
\begin{aligned}
&|D(x_{\text{FN}, 1}^{t-1,\text{DI}})-D(x_{\text{FN}, 0}^{t-1,\text{DI}})|
\\
= & \alpha  \left|\frac{\|\nabla_{x_{\text{FN}, 1}^{t-1,\text{DI}}}f(x_{\text{FN}, 1}^{t-1,\text{DI}})\|_1}{f(x_{\text{FN}, 1}^{t-1,\text{DI}})}  - \frac{\|\nabla_{x_{\text{FN}, 0}^{t-1,\text{DI}}}f(x_{\text{FN}, 0}^{t-1,\text{DI}})\|_1}{f(x_{\text{FN}, 0}^{t-1,\text{DI}})} \right|
\\
\le &  \alpha\left|\frac{\sqrt{n}Kd(x_{\text{FN}, 1}^{t-1,\text{DI}},x_{\text{FN}, 0}^{t-1,\text{DI}}) +  \|\nabla_{x_{\text{FN}, 0}^{t-1,\text{DI}}}f(x_{\text{FN}, 0}^{t-1,\text{DI}})\|_1}{f(x_{\text{FN}, 1}^{t-1,\text{DI}})}  - \frac{\|\nabla_{x_{\text{FN}, 0}^{t-1,\text{DI}}}f(x_{\text{FN}, 0}^{t-1,\text{DI}})\|_1}{f(x_{\text{FN}, 0}^{t-1,\text{DI}})} \right|
\\
\le &  \frac{\sqrt{n}\alpha Kd(x_{\text{FN}, 1}^{t-1,\text{DI}},x_{\text{FN}, 0}^{t-1,\text{DI}})}{f(x_{\text{FN}, 1}^{t-1,\text{DI}})} + \left|\frac{ \alpha \|\nabla_{x_{\text{FN}, 0}^{t-1,\text{DI}}}f(x_{\text{FN}, 0}^{t-1,\text{DI}})\|_1}{f(x_{\text{FN}, 1}^{t-1,\text{DI}})}  - \frac{ \alpha\|\nabla_{x_{\text{FN}, 0}^{t-1,\text{DI}}}f(x_{\text{FN}, 0}^{t-1,\text{DI}})\|_1}{f(x_{\text{FN}, 0}^{t-1,\text{DI}})} \right|,
\end{aligned}
\end{equation}
where $d(x,y) \coloneqq \|x-y\|_2$ is the distance between the two feature. Taking the summation over $T$ iterations, we have
\begin{equation}
|D(x_{\text{FN},1})-D(x_{\text{FN},0})| \le \sum_{t=1}^{T} \left[\frac{\sqrt{n}\alpha Kd(x_{\text{FN}, 1}^{t-1,\text{DI}},x_{\text{FN}, 0}^{t-1,\text{DI}})}{f(x_{\text{FN}, 1}^{t-1,\text{DI}})} + \alpha \left|\frac{ f(x_{\text{FN}, 0}^{t-1,\text{DI}}) - f(x_{\text{FN}, 1}^{t-1,\text{DI}}) }{f(x_{\text{FN}, 1}^{t-1,\text{DI}})f(x_{\text{FN}, 0}^{t-1,\text{DI}})} \right|\delta_{\text{FN},0}^{t-1}\right]. 
\label{acr-eod-upper}
\end{equation}

Since the above inequality holds true for all disadvantaged TP samples, we can further write \eqref{acr-eod-upper} as
\begin{equation*}
|D(x_{\text{FN},1})-D(x_{\text{FN},0})| \le \min_{x_{\text{FN},0} \in \mathbb{S}_{10}}\sum_{t=1}^{T} \left[\frac{\sqrt{n}\alpha Kd(x_{\text{FN}, 1}^{t-1,\text{DI}},x_{\text{FN}, 0}^{t-1,\text{DI}})}{f(x_{\text{FN}, 1}^{t-1,\text{DI}})} + \alpha \left|\frac{ f(x_{\text{FN}, 0}^{t-1,\text{DI}}) - f(x_{\text{FN}, 1}^{t-1,\text{DI}}) }{f(x_{\text{FN}, 1}^{t-1,\text{DI}})f(x_{\text{FN}, 0}^{t-1,\text{DI}})} \right|\delta_{\text{FN},0}^{t-1}\right],
\end{equation*}
where $\delta_{\text{TP},0}^{t-1} \coloneqq \delta \|\nabla_{x_{\text{FN}, 0}^{t-1,\text{DI}}}f_{\theta}(x_{\text{FN}, 0}^{t-1,\text{DI}})\|_1$ is the change of $x_{\text{FN},0}$'s predicted label under $\epsilon$-level accuracy attack at $t-1$-th iteration. This shows that under DI attack, the difference of change in performance regarding marginal advantaged FN samples are upper-bounded by the robustness of marginal disadvantaged FN samples up to an addictive constant. For $f$ under normal training and $f'$ under normal training, we have similar upper-bound except that we now have $\delta_{\text{FN},0}^{'^{t-1}} \ge \delta_{\text{FN},0}^{t-1}$, which indicates that the adversarial classifier achieves tighter upper-bound than that of a normal classifier. For $\frac{\|\nabla_{x_{\text{FN}, 1}^{t-1,\text{DI}}}f(x_{\text{FN}, 1}^{t-1,\text{DI}})\|_1}{f(x_{\text{FN}, 1}^{t-1,\text{DI}})} \le \frac{\|\nabla_{x_{\text{FN}, 0}^{t-1,\text{DI}}}f(x_{\text{FN}, 0}^{t-1,\text{DI}})\|_1}{f(x_{\text{FN}, 0}^{t-1,\text{DI}})}$, we have same upper-bound:
\begin{equation*}
\begin{aligned}
&|D(x_{\text{FN},1}^t)-D(x_{\text{FN},0}^t)|
\\
= & \alpha\left|\frac{\|\nabla_{x_{\text{FN}, 1}^{t-1,\text{DI}}}f(x_{\text{FN}, 1}^{t-1,\text{DI}})\|_1}{f(x_{\text{FN}, 1}^{t-1,\text{DI}})} -\frac{\|\nabla_{x_{\text{FN}, 0}^{t-1,\text{DI}}}f(x_{\text{FN}, 0}^{t-1,\text{DI}})\|_1}{f(x_{\text{FN}, 0}^{t-1,\text{DI}})}\right|
\\
\le & \alpha\left|\frac{\|\nabla_{x_{\text{FN}, 0}^{t-1,\text{DI}}}f(x_{\text{FN}, 0}^{t-1,\text{DI}})\|_1}{f(x_{\text{FN}, 0}^{t-1,\text{DI}})} - \frac{\|\nabla_{x_{\text{FN}, 0}^{t-1,\text{DI}}}f(x_{\text{FN}, 0}^{t-1,\text{DI}})\|_1-\sqrt{n}Kd(x_{\text{FN}, 1}^{t-1,\text{DI}},x_{\text{FN}, 0}^{t-1,\text{DI}})}{f(x_{\text{FN}, 0}^{t-1,\text{DI}})}\right|
\\
\le &  \frac{\sqrt{n}\alpha Kd(x_{\text{FN}, 1}^{t-1,\text{DI}},x_{\text{FN}, 0}^{t-1,\text{DI}})}{f(x_{\text{FN}, 1}^{t-1,\text{DI}})} + \left|\frac{ \alpha \|\nabla_{x_{\text{FN}, 0}^{t-1,\text{DI}}}f(x_{\text{FN}, 0}^{t-1,\text{DI}})\|_1}{f(x_{\text{FN}, 1}^{t-1,\text{DI}})}  - \frac{ \alpha\|\nabla_{x_{\text{FN}, 0}^{t-1,\text{DI}}}f(x_{\text{FN}, 0}^{t-1,\text{DI}})\|_1}{f(x_{\text{FN}, 0}^{t-1,\text{DI}})} \right|.
\end{aligned}
\end{equation*}
\end{proof}

\subsection{Proof of Theorem \ref{eod-acr}}

\begin{proof}
Let $f$ be the function of classifier, consider $x_{\text{TP},0}$, we have the predicted soft label for sample $x_{\text{TP},0}$ under accuracy attack at $t-1$-th iteration as follows:
\begin{equation*}
\begin{aligned}
&f(x^{t,\text{Acc}}_{\text{TP},0})
\\
=&f(x^{t-1,\text{Acc}}_{\text{TP},0} + \alpha\text{sign}(\nabla_{x^{t-1,\text{Acc}}_{\text{TP},0}}L))
\\
\approx &f(x^{t-1,\text{Acc}}_{\text{TP},0}) + \alpha (\nabla_{x^{t-1,\text{Acc}}_{\text{TP},0}} f(x^{t-1,\text{Acc}}_{\text{TP},0}))^T \text{sign}(-\frac{1}{f(x^{t-1,\text{Acc}}_{\text{TP},0})}\nabla_{x^{t-1,\text{Acc}}_{\text{TP},0}} f(x^{t-1,\text{Acc}}_{\text{TP},0}))
\\
= &f(x^{t-1,\text{Acc}}_{\text{TP},0}) + \alpha (\nabla_{x} f(x^{t-1,\text{Acc}}_{\text{TP},0}))^T \text{sign}(\nabla_{x^{t-1,\text{Acc}}_{\text{TP},0}}L)
\\
=& f(x^{t-1,\text{Acc}}_{\text{TP},0}) -\alpha \|\nabla_{x^{t-1,\text{Acc}}_{\text{TP},0}}f(x^{t-1,\text{Acc}}_{\text{TP},0})\|_1 
\\
= & f(x^{t-1,\text{Acc}}_{\text{TP},0}) - \xi_{\text{TP},0}^{t-1},
\end{aligned}
\end{equation*}

where $\xi_{\text{TP},0}^{t-1} \coloneqq \alpha \|\nabla_{x^{t-1,\text{Acc}}_{\text{TP},0}}f(x^{t-1,\text{Acc}}_{\text{TP},0})\|_1$ is the change of $x_{\text{TP},0}$'s predicted label under $\epsilon$-level DI or accuracy attack at $t-1$-th iteration since both are equivalent regarding $x_{\text{TP},0}$. This shows that disadvantaged TP samples that attains $\delta$-level robustness under $\epsilon$-level DI attack also attains similar robustness w.r.t. accuracy attack.

For $x_{\text{TP},1}$, let $F(x^{t,\text{Acc}}_{\text{TP},1}):=|f(x^{t,\text{Acc}}_{\text{TP},1}) -f(x^{t-1,\text{Acc}}_{\text{TP},1})|$, we have its change in predicted soft label under accuracy attack at $t-1$-th iteration as follows:
\begin{equation}
\begin{aligned}
&F(x^{t,\text{Acc}}_{\text{TP},1})
\\
=&|f(x^{t,\text{Acc}}_{\text{TP},1}) -f(x^{t-1,\text{Acc}}_{\text{TP},1})|
\\
=&|f(x^{t-1,\text{Acc}}_{\text{TP},1} + \alpha\text{sign}(\nabla_{x^{t-1,\text{Acc}}_{\text{TP},1}}L)) -f(x^{t-1,\text{Acc}}_{\text{TP},1})|
\\
\approx & \alpha (\nabla_{x^{t-1,\text{Acc}}_{\text{TP},1}} f(x^{t-1,\text{Acc}}_{\text{TP},1}))^T \text{sign}(\nabla_{x^{t-1,\text{Acc}}_{\text{TP},1}}L_\text{CE})
\\
=& \alpha \|\nabla_{x^{t-1,\text{Acc}}_{\text{TP},1}} f(x^{t-1,\text{Acc}}_{\text{TP},1})\|_1
\\
\le & \xi_{\text{TP},0}^{t-1} +\sqrt{n}\alpha Kd(x^{t-1,\text{Acc}}_{\text{TP},0},x^{t-1,\text{Acc}}_{\text{TP},1}).
\end{aligned}
\end{equation}

Taking the summation over all iterations, we have

\begin{equation}
F(x_{\text{TP},1}) \le \xi_{\text{TP},0} +\sum_{t=1}^T \sqrt{n}\alpha Kd(x^{t-1,\text{Acc}}_{\text{TP},0},x^{t-1,\text{Acc}}_{\text{TP},1}),
\label{eod-acr-upper}
\end{equation}
where $\xi_{\text{TP},0}$ is the change of predicted soft label of sample $x_{\text{TP},0}$ under $\epsilon$-level PGD attack. Since the inequality hold true for all $x_{\text{TP},0}$, we can further write \eqref{eod-acr-upper} as
\begin{equation*}
F(x_{\text{TP},1}) \le \min_{x_{\text{TP},0} \in \mathbb{S}_{10}} \xi_{\text{TP},0} +\sum_{t=1}^T \sqrt{n}\alpha Kd(x^{t-1,\text{Acc}}_{\text{TP},0},x^{t-1,\text{Acc}}_{\text{TP},1}).
\end{equation*}

And the lower bound $F(x_{\text{TP},1}) \geq 0$ naturally holds true for samples under accuracy attack. 
This shows that for samples in the advantaged group, the change of predicted soft label under accuracy attack is lower-bounded by the robustness of its neighbor sample(s) in the disadvantaged group up to an addictive constant. For $f$ under fairness adversarial training and $f'$ under normal training, we have similar upper-bound except that we now have $\xi'_{\text{TP},0} \ge \xi_{\text{TP},0}$, which indicates that the adversarial classifier achieves tighter upper-bound than that of a normal classifier. 
\end{proof}

\subsection{results of robustness against DI attack}

We include the results of fair adversarial training in Tab. \ref{acr-adt}-\ref{otnr-celeba} to better distinguish between different fairness methods.

\subsection{more results on robustness against accuracy attack}

\begin{figure*}[!h]
    \centering
    \subfigure[Accuracy]{\includegraphics[width=0.23\linewidth]{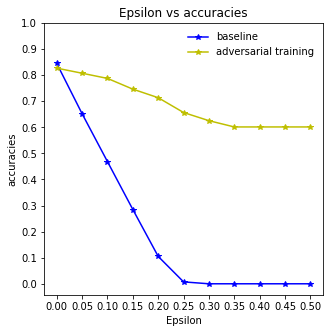}}
    \subfigure[DI]{\includegraphics[width=0.23\linewidth]{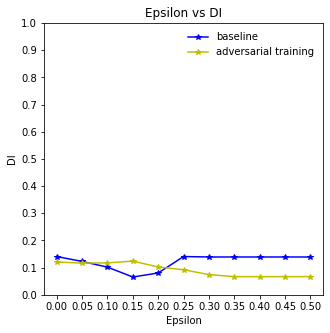}}
    \subfigure[EOd]{\includegraphics[width=0.23\linewidth]{adv_eod_cps_d2.png}}
\caption{Results of a classifier adversarially trained w.r.t. DI. Change of accuracy, DI and EOd under accuracy attack on COMPAS dataset.}
    \label{acr_adv_cps}
\end{figure*}

\begin{figure*}[!h]
    \centering
    \subfigure[Accuracy]{\includegraphics[width=0.23\linewidth]{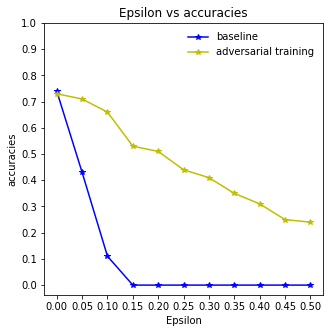}}
    \subfigure[DI]{\includegraphics[width=0.23\linewidth]{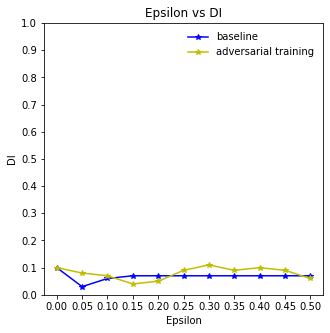}}
    \subfigure[EOd]{\includegraphics[width=0.23\linewidth]{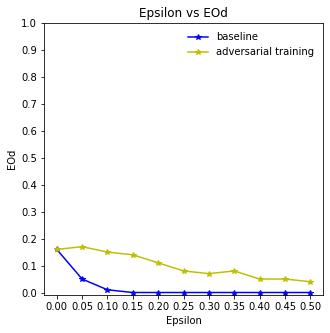}}
\caption{Results of a classifier adversarially trained w.r.t. DI. Change of accuracy, DI and EOd under accuracy attack on German dataset.}
    \label{acr_adv_german}
\end{figure*}

\begin{figure*}[!h]
    \centering
    \subfigure[Accuracy]{\includegraphics[width=0.24\linewidth]{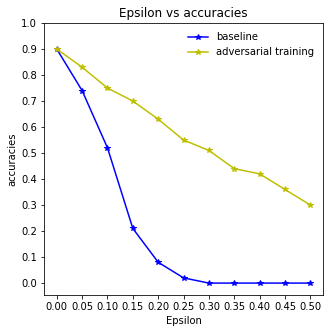}}
    \subfigure[DI]{\includegraphics[width=0.24\linewidth]{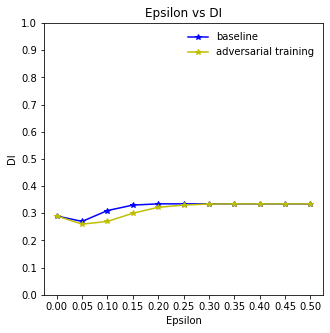}}
    \subfigure[EOd]{\includegraphics[width=0.24\linewidth]{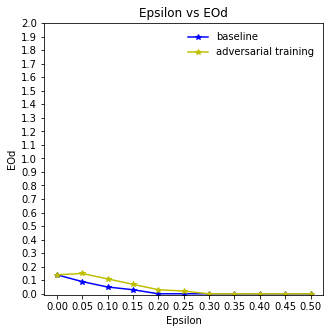}}
\caption{Results of a classifier adversarially trained w.r.t. DI. Change of accuracy, DI and EOd under accuracy attack on CelebA dataset.}
    \label{acr_adv_celeba}
\end{figure*}

We show the results on robustness against accuracy attack on COMPAS, GERMAN and CelebA datasets in Fig. \ref{acr_adv_cps}-\ref{acr_adv_celeba}.

\begin{table*}[!h]
\centering
\begin{tabular}{llll}
\hline
M  &adv+pre & adv+in & adv+post\\
\hline
0.000 &      0.800 &      0.800 &      0.800 \\
0.050 &      0.790 &      0.800 &      0.790 \\
0.100 &      0.795 &      0.795 &      0.790 \\
0.150 &      0.794 &      0.794 &      0.790 \\
0.200 &      0.794 &      0.794 &      0.790 \\
0.250 &      0.784 &      0.794 &      0.784 \\
0.300 &      0.788 &      0.788 &      0.781 \\
0.350 &      0.771 &      0.781 &      0.771 \\
0.400 &      0.778 &      0.778 &      0.771 \\
0.450 &      0.776 &      0.776 &      0.774 \\
0.500 &      0.771 &      0.771 &      0.772 \\
\hline
\end{tabular}
\caption{results of accuracy for adversarial fair training on Adult dataset under DI attack.}
\label{acr-adt}
\end{table*}

\begin{table*}[!h]
\centering
\begin{tabular}{llll}
\hline
M  &adv+pre & adv+in & adv+post\\
\hline
0.000 &      0.029 &      0.039 &      0.016 \\
0.050 &      0.117 &      0.137 &      0.098 \\
0.100 &      0.129 &      0.111 &      0.119 \\
0.150 &      0.128 &      0.108 &      0.108 \\
0.200 &      0.114 &      0.104 &      0.114 \\
0.250 &      0.123 &      0.093 &      0.123 \\
0.300 &      0.114 &      0.074 &      0.104 \\
0.350 &      0.099 &      0.059 &      0.090 \\
0.400 &      0.086 &      0.046 &      0.096 \\
0.450 &      0.103 &      0.073 &      0.113 \\
0.500 &      0.152 &      0.152 &      0.132 \\
\hline
\end{tabular}
\caption{results of EOd for adversarial fair training on Adult dataset under DI attack.}
\end{table*}

\begin{table*}[!h]
\centering
\begin{tabular}{llll}
\hline
M  &adv+pre & adv+in & adv+post\\
\hline
0.000 &      0.050 &      0.050 &      0.050 \\
0.050 &      0.067 &      0.067 &      0.067 \\
0.100 &      0.066 &      0.056 &      0.063 \\
0.150 &      0.066 &      0.054 &      0.066 \\
0.200 &      0.070 &      0.050 &      0.070 \\
0.250 &      0.077 &      0.047 &      0.072 \\
0.300 &      0.068 &      0.043 &      0.068 \\
0.350 &      0.080 &      0.040 &      0.087 \\
0.400 &      0.090 &      0.040 &      0.090 \\
0.450 &      0.087 &      0.047 &      0.087 \\
0.500 &      0.088 &      0.058 &      0.083 \\
\hline
\end{tabular}
\caption{results of DI for adversarial fair training on Adult dataset under DI attack.}
\end{table*}

\begin{table*}[!h]
\centering
\begin{tabular}{llll}
\hline
M  &adv+pre & adv+in & adv+post\\
\hline
0.000 &      0.268 &      0.275 &      0.282 \\
0.050 &      0.286 &      0.286 &      0.286 \\
0.100 &      0.243 &      0.243 &      0.246 \\
0.150 &      0.235 &      0.231 &      0.235 \\
0.200 &      0.227 &      0.226 &      0.227 \\
0.244 &      0.225 &      0.225 &      0.225 \\
0.300 &      0.205 &      0.205 &      0.211 \\
0.350 &      0.183 &      0.188 &      0.186 \\
0.400 &      0.184 &      0.184 &      0.181 \\
0.450 &      0.195 &      0.193 &      0.193 \\
0.500 &      0.203 &      0.203 &      0.207 \\
\hline
\end{tabular}
\caption{results of white TPR for adversarial fair training on Adult dataset under DI attack.}
\end{table*}

\begin{table*}[!h]
\centering
\begin{tabular}{llll}
\hline
M  &adv+pre & adv+in & adv+post\\
\hline
0.000 &      0.973 &      0.973 &      0.973 \\
0.050 &      0.970 &      0.970 &      0.970 \\
0.100 &      0.977 &      0.977 &      0.977 \\
0.150 &      0.979 &      0.979 &      0.979 \\
0.200 &      0.982 &      0.982 &      0.982 \\
0.250 &      0.983 &      0.983 &      0.983 \\
0.300 &      0.981 &      0.981 &      0.981 \\
0.350 &      0.967 &      0.977 &      0.967 \\
0.400 &      0.964 &      0.974 &      0.964 \\
0.450 &      0.957 &      0.967 &      0.957 \\
0.500 &      0.958 &      0.958 &      0.958 \\
\hline
\end{tabular}
\caption{results of white TNR for adversarial fair training on Adult dataset under DI attack.}
\end{table*}

\begin{table*}[!h]
\centering
\begin{tabular}{llll}
\hline
M  &adv+pre & adv+in & adv+post\\
\hline
0.000 &      0.268 &      0.248 &      0.262 \\
0.050 &      0.208 &      0.168 &      0.201 \\
0.100 &      0.168 &      0.148 &      0.168 \\
0.150 &      0.141 &      0.141 &      0.151 \\
0.200 &      0.134 &      0.134 &      0.134 \\
0.250 &      0.141 &      0.141 &      0.141 \\
0.300 &      0.131 &      0.144 &      0.135 \\
0.350 &      0.141 &      0.140 &      0.143 \\
0.400 &      0.154 &      0.151 &      0.158 \\
0.450 &      0.141 &      0.141 &      0.143 \\
0.500 &      0.074 &      0.074 &      0.094 \\
\hline
\end{tabular}
\caption{results of black TPR for adversarial fair training on Adult dataset under DI attack.}
\end{table*}

\begin{table*}[!h]
\centering
\begin{tabular}{llll}
\hline
M  &adv+pre & adv+in & adv+post\\
\hline
0.000 &      0.989 &      0.989 &      0.984 \\
0.050 &      0.989 &      0.981 &      0.987 \\
0.100 &      0.990 &      0.992 &      0.992 \\
0.150 &      0.993 &      0.997 &      0.995 \\
0.200 &      0.993 &      0.993 &      0.993 \\
0.250 &      0.991 &      0.991 &      0.991 \\
0.300 &      0.990 &      0.986 &      0.991 \\
0.350 &      0.991 &      0.993 &      0.990 \\
0.400 &      0.986 &      0.990 &      0.990 \\
0.450 &      0.988 &      0.984 &      0.988 \\
0.500 &      0.978 &      0.982 &      0.976 \\
\hline
\end{tabular}
\caption{results of black TNR for adversarial fair training on Adult dataset under DI attack.}
\end{table*}

\begin{table*}[!h]
\centering
\begin{tabular}{llll}
\hline
M  &adv+pre & adv+in & adv+post\\
\hline
0.000 &      0.625 &      0.627 &      0.635 \\
0.010 &      0.624 &      0.609 &      0.634 \\
0.050 &      0.617 &      0.601 &      0.627 \\
0.100 &      0.607 &      0.607 &      0.604 \\
0.150 &      0.603 &      0.610 &      0.603 \\
0.200 &      0.607 &      0.610 &      0.602 \\
0.250 &      0.612 &      0.606 &      0.612 \\
0.300 &      0.603 &      0.592 &      0.603 \\
0.350 &      0.598 &      0.579 &      0.598 \\
0.400 &      0.595 &      0.567 &      0.595 \\
0.450 &      0.588 &      0.558 &      0.588 \\
0.500 &      0.586 &      0.551 &      0.586 \\
\hline
\end{tabular}
\caption{results of accuracy for adversarial fair training on COMPAS dataset under DI attack.}
\end{table*}

\begin{table*}[!h]
\centering
\begin{tabular}{llll}
\hline
M  &adv+pre & adv+in & adv+post\\
\hline
0.000 &      0.044 &      0.240 &      0.024 \\
0.010 &      0.197 &      0.584 &      0.147 \\
0.050 &      0.735 &      0.979 &      0.565 \\
0.100 &      0.960 &      1.146 &      0.910 \\
0.150 &      1.131 &      1.231 &      1.041 \\
0.200 &      1.214 &      1.289 &      1.254 \\
0.250 &      1.348 &      1.387 &      1.348 \\
0.300 &      1.463 &      1.502 &      1.463 \\
0.350 &      1.513 &      1.598 &      1.513 \\
0.400 &      1.623 &      1.645 &      1.623 \\
0.450 &      1.676 &      1.665 &      1.676 \\
0.500 &      1.705 &      1.710 &      1.705 \\
\hline
\end{tabular}
\caption{results of EOd for adversarial fair training on COMPAS dataset under DI attack.}
\end{table*}

\begin{table*}[!h]
\centering
\begin{tabular}{llll}
\hline
M  &adv+pre & adv+in & adv+post\\
\hline
0.000 &      0.070 &      0.133 &      0.050 \\
0.010 &      0.154 &      0.302 &      0.134 \\
0.050 &      0.317 &      0.488 &      0.297 \\
0.100 &      0.396 &      0.572 &      0.356 \\
0.150 &      0.471 &      0.614 &      0.471 \\
0.200 &      0.588 &      0.643 &      0.588 \\
0.250 &      0.645 &      0.692 &      0.645 \\
0.300 &      0.716 &      0.750 &      0.716 \\
0.350 &      0.765 &      0.798 &      0.765 \\
0.400 &      0.809 &      0.822 &      0.809 \\
0.450 &      0.830 &      0.832 &      0.830 \\
0.500 &      0.844 &      0.855 &      0.844 \\
\hline
\end{tabular}
\caption{results of DI for adversarial fair training on COMPAS dataset under DI attack.}
\end{table*}

\begin{table*}[!h]
\centering
\begin{tabular}{llll}
\hline
M  &adv+pre & adv+in & adv+post\\
\hline
0.000 &      0.311 &      0.336 &      0.321 \\
0.010 &      0.267 &      0.229 &      0.297 \\
0.050 &      0.072 &      0.014 &      0.172 \\
0.100 &      0.000 &      0.000 &      0.021 \\
0.150 &      0.000 &      0.000 &      0.003 \\
0.200 &      0.000 &      0.000 &      0.000 \\
0.250 &      0.000 &      0.000 &      0.000 \\
0.300 &      0.000 &      0.000 &      0.000 \\
0.350 &      0.000 &      0.000 &      0.000 \\
0.400 &      0.000 &      0.000 &      0.000 \\
0.450 &      0.000 &      0.000 &      0.000 \\
0.500 &      0.000 &      0.000 &      0.000 \\
\hline
\end{tabular}
\caption{results of white TPR for adversarial fair training on COMPAS dataset under DI attack.}
\end{table*}

\begin{table*}[!h]
\centering
\begin{tabular}{llll}
\hline
M  &adv+pre & adv+in & adv+post\\
\hline
0.000 &      0.914 &      0.788 &      0.914 \\
0.010 &      0.935 &      0.864 &      0.935 \\
0.050 &      1.000 &      0.983 &      1.000 \\
0.100 &      1.000 &      1.000 &      1.000 \\
0.150 &      1.000 &      1.000 &      1.000 \\
0.200 &      1.000 &      1.000 &      1.000 \\
0.250 &      1.000 &      1.000 &      1.000 \\
0.300 &      1.000 &      1.000 &      1.000 \\
0.350 &      1.000 &      1.000 &      1.000 \\
0.400 &      1.000 &      1.000 &      1.000 \\
0.450 &      1.000 &      1.000 &      1.000 \\
0.500 &      1.000 &      1.000 &      1.000 \\
\hline
\end{tabular}
\caption{results of white TNR for adversarial fair training on COMPAS dataset under DI attack.}
\end{table*}

\begin{table*}[!h]
\centering
\begin{tabular}{llll}
\hline
M  &adv+pre & adv+in & adv+post\\
\hline
0.000 &      0.339 &      0.525 &      0.339 \\
0.010 &      0.385 &      0.573 &      0.365 \\
0.050 &      0.565 &      0.596 &      0.565 \\
0.100 &      0.599 &      0.672 &      0.599 \\
0.150 &      0.635 &      0.720 &      0.635 \\
0.200 &      0.695 &      0.749 &      0.695 \\
0.250 &      0.760 &      0.793 &      0.760 \\
0.300 &      0.808 &      0.828 &      0.808 \\
0.350 &      0.836 &      0.858 &      0.836 \\
0.400 &      0.868 &      0.862 &      0.868 \\
0.450 &      0.890 &      0.858 &      0.890 \\
0.500 &      0.894 &      0.870 &      0.894 \\
\hline
\end{tabular}
\caption{results of black TPR for adversarial fair training on COMPAS dataset under DI attack.}
\end{table*}

\begin{table*}[!h]
\centering
\begin{tabular}{llll}
\hline
M  &adv+pre & adv+in & adv+post\\
\hline
0.000 &      0.908 &      0.736 &      0.908 \\
0.010 &      0.866 &      0.625 &      0.886 \\
0.050 &      0.748 &      0.586 &      0.798 \\
0.100 &      0.559 &      0.525 &      0.559 \\
0.150 &      0.524 &      0.489 &      0.524 \\
0.200 &      0.471 &      0.460 &      0.471 \\
0.250 &      0.401 &      0.406 &      0.401 \\
0.300 &      0.325 &      0.327 &      0.325 \\
0.350 &      0.253 &      0.260 &      0.253 \\
0.400 &      0.195 &      0.217 &      0.195 \\
0.450 &      0.174 &      0.193 &      0.174 \\
0.500 &      0.148 &      0.160 &      0.148 \\
\hline
\end{tabular}
\caption{results of black TNR for adversarial fair training on COMPAS dataset under DI attack.}
\end{table*}

\begin{table*}[!h]
\centering
\begin{tabular}{llll}
\hline
M  &adv+pre & adv+in & adv+post\\
\hline
0.000 &      0.724 &      0.714 &      0.730 \\
0.050 &      0.721 &      0.711 &      0.726 \\
0.100 &      0.710 &      0.700 &      0.721 \\
0.150 &      0.690 &      0.690 &      0.714 \\
0.200 &      0.680 &      0.680 &      0.703 \\
0.250 &      0.684 &      0.684 &      0.690 \\
0.300 &      0.680 &      0.680 &      0.680 \\
0.350 &      0.676 &      0.670 &      0.676 \\
0.400 &      0.667 &      0.667 &      0.667 \\
0.450 &      0.665 &      0.665 &      0.665 \\
0.500 &      0.660 &      0.667 &      0.665 \\
\hline
\end{tabular}
\caption{results of accuracy for adversarial fair training on German dataset under DI attack.}
\end{table*}

\begin{table*}[!h]
\centering
\begin{tabular}{llll}
\hline
M  &adv+pre & adv+in & adv+post\\
\hline
0.000 &      0.030 &      0.030 &      0.010 \\
0.050 &      0.140 &      0.140 &      0.100 \\
0.100 &      0.380 &      0.380 &      0.380 \\
0.150 &      0.770 &      0.770 &      0.670 \\
0.200 &      0.960 &      0.960 &      0.950 \\
0.250 &      1.270 &      1.270 &      1.250 \\
0.300 &      1.290 &      1.340 &      1.290 \\
0.350 &      1.310 &      1.360 &      1.330 \\
0.400 &      1.340 &      1.460 &      1.340 \\
0.450 &      1.440 &      1.540 &      1.490 \\
0.500 &      1.500 &      1.600 &      1.560 \\
\hline
\end{tabular}
\caption{results of EOd for adversarial fair training on German dataset under DI attack.}
\end{table*}

\begin{table*}[!h]
\centering
\begin{tabular}{llll}
\hline
M  &adv+pre & adv+in & adv+post\\
\hline
0.000 &      0.060 &      0.120 &      0.020 \\
0.050 &      0.130 &      0.130 &      0.110 \\
0.100 &      0.170 &      0.170 &      0.140 \\
0.150 &      0.260 &      0.260 &      0.270 \\
0.200 &      0.300 &      0.300 &      0.340 \\
0.250 &      0.360 &      0.360 &      0.360 \\
0.300 &      0.410 &      0.410 &      0.440 \\
0.350 &      0.470 &      0.470 &      0.470 \\
0.400 &      0.580 &      0.580 &      0.550 \\
0.450 &      0.640 &      0.640 &      0.600 \\
0.500 &      0.670 &      0.670 &      0.710 \\
\hline
\end{tabular}
\caption{results of DI for adversarial fair training on German dataset under DI attack.}
\end{table*}

\begin{table*}[!h]
\centering
\begin{tabular}{llll}
\hline
M  &adv+pre & adv+in & adv+post\\
\hline
0.000 &      0.364 &      0.364 &      0.364 \\
0.050 &      0.350 &      0.350 &      0.350 \\
0.100 &      0.260 &      0.260 &      0.260 \\
0.150 &      0.130 &      0.130 &      0.190 \\
0.200 &      0.110 &      0.110 &      0.110 \\
0.250 &      0.070 &      0.070 &      0.070 \\
0.300 &      0.000 &      0.000 &      0.000 \\
0.350 &      0.000 &      0.000 &      0.000 \\
0.400 &      0.000 &      0.000 &      0.000 \\
0.450 &      0.000 &      0.000 &      0.000 \\
0.500 &      0.000 &      0.000 &      0.000 \\
\hline
\end{tabular}
\caption{results of male TPR for adversarial fair training on German dataset under DI attack.}
\end{table*}

\begin{table*}[!h]
\centering
\begin{tabular}{llll}
\hline
M  &adv+pre & adv+in & adv+post\\
\hline
0.000 &      0.857 &      0.857 &      0.850 \\
0.050 &      0.870 &      0.870 &      0.870 \\
0.100 &      0.920 &      0.920 &      0.920 \\
0.150 &      1.000 &      1.000 &      1.000 \\
0.200 &      1.000 &      1.000 &      1.000 \\
0.250 &      1.000 &      1.000 &      1.000 \\
0.300 &      1.000 &      1.000 &      1.000 \\
0.350 &      1.000 &      1.000 &      1.000 \\
0.400 &      1.000 &      1.000 &      1.000 \\
0.450 &      1.000 &      1.000 &      1.000 \\
0.500 &      1.000 &      1.000 &      1.000 \\
\hline
\end{tabular}
\caption{results of male TNR for adversarial fair training on German dataset under DI attack.}
\end{table*}

\begin{table*}[!h]
\centering
\begin{tabular}{llll}
\hline
M  &adv+pre & adv+in & adv+post\\
\hline
0.000 &      0.377 &      0.377 &      0.377 \\
0.050 &      0.420 &      0.420 &      0.420 \\
0.100 &      0.510 &      0.510 &      0.510 \\
0.150 &      0.570 &      0.570 &      0.570 \\
0.200 &      0.680 &      0.680 &      0.680 \\
0.250 &      0.750 &      0.750 &      0.750 \\
0.300 &      0.770 &      0.770 &      0.770 \\
0.350 &      0.780 &      0.780 &      0.790 \\
0.400 &      0.800 &      0.810 &      0.800 \\
0.450 &      0.860 &      0.870 &      0.860 \\
0.500 &      0.870 &      0.890 &      0.870 \\
\hline
\end{tabular}
\caption{results of female TPR for adversarial fair training on German dataset under DI attack.}
\end{table*}

\begin{table*}[!h]
\centering
\begin{tabular}{llll}
\hline
M  &adv+pre & adv+in & adv+post\\
\hline
0.000 &      0.833 &      0.833 &      0.843 \\
0.050 &      0.810 &      0.810 &      0.820 \\
0.100 &      0.740 &      0.740 &      0.740 \\
0.150 &      0.670 &      0.670 &      0.690 \\
0.200 &      0.560 &      0.560 &      0.560 \\
0.250 &      0.490 &      0.490 &      0.510 \\
0.300 &      0.480 &      0.440 &      0.480 \\
0.350 &      0.460 &      0.410 &      0.460 \\
0.400 &      0.450 &      0.340 &      0.450 \\
0.450 &      0.400 &      0.310 &      0.370 \\
0.500 &      0.300 &      0.240 &      0.230 \\
\hline
\end{tabular}
\caption{results of female TNR for adversarial fair training on German dataset under DI attack.}
\label{ftnr-german}
\end{table*}

\begin{table*}[!h]
\centering
\begin{tabular}{llll}
\hline
M  &adv+pre & adv+in & adv+post\\
\hline
0.000 & 0.870 & 0.840 & 0.860 \\
0.050 & 0.850 & 0.830 & 0.840 \\
0.100 & 0.830 & 0.810 & 0.810 \\
0.150 & 0.810 & 0.790 & 0.800 \\
0.200 & 0.760 & 0.770 & 0.760 \\
0.250 & 0.740 & 0.740 & 0.740 \\
0.300 & 0.700 & 0.710 & 0.710 \\
0.350 & 0.680 & 0.690 & 0.680 \\
0.400 & 0.670 & 0.650 & 0.660 \\
0.450 & 0.640 & 0.630 & 0.620 \\
0.500 & 0.620 & 0.600 & 0.590 \\
\hline
\end{tabular}
\caption{results of accuracy for adversarial fair training on CelebA dataset under DI attack.}
\end{table*}

\begin{table*}[!h]
\centering
\begin{tabular}{llll}
\hline
M  &adv+pre & adv+in & adv+post\\
\hline
0.000 & 0.030 & 0.030 & 0.040 \\
0.050 & 0.110 & 0.110 & 0.130 \\
0.100 & 0.200 & 0.230 & 0.250 \\
0.150 & 0.300 & 0.320 & 0.310 \\
0.200 & 0.390 & 0.440 & 0.380 \\
0.250 & 0.490 & 0.490 & 0.470 \\
0.300 & 0.550 & 0.550 & 0.520 \\
0.350 & 0.600 & 0.640 & 0.600 \\
0.400 & 0.650 & 0.660 & 0.660 \\
0.450 & 0.700 & 0.710 & 0.690 \\
0.500 & 0.720 & 0.780 & 0.730 \\
\hline
\end{tabular}
\caption{results of EOd for adversarial fair training on CelebA dataset under DI attack.}
\end{table*}

\begin{table*}[!h]
\centering
\begin{tabular}{llll}
\hline
M  &adv+pre & adv+in & adv+post\\
\hline
0.000 & 0.050 & 0.070 & 0.040 \\
0.050 & 0.090 & 0.110 & 0.110 \\
0.100 & 0.160 & 0.114 & 0.130 \\
0.150 & 0.230 & 0.210 & 0.210 \\
0.200 & 0.260 & 0.290 & 0.280 \\
0.250 & 0.300 & 0.330 & 0.340 \\
0.300 & 0.340 & 0.360 & 0.390 \\
0.350 & 0.390 & 0.380 & 0.410 \\
0.400 & 0.400 & 0.420 & 0.440 \\
0.450 & 0.410 & 0.440 & 0.450 \\
0.500 & 0.440 & 0.450 & 0.470 \\
\hline
\end{tabular}
\caption{results of DI for adversarial fair training on CelebA dataset under DI attack.}
\end{table*}

\begin{table*}[!h]
\centering
\begin{tabular}{llll}
\hline
M  &adv+pre & adv+in & adv+post\\
\hline
0.000 & 0.830 & 0.840 & 0.850 \\
0.050 & 0.880 & 0.860 & 0.870 \\
0.100 & 0.920 & 0.900 & 0.940 \\
0.150 & 0.970 & 0.950 & 0.960 \\
0.200 & 0.990 & 0.990 & 1.000 \\
0.250 & 1.000 & 1.000 & 1.000 \\
0.300 & 1.000 & 1.000 & 1.000 \\
0.350 & 1.000 & 1.000 & 1.000 \\
0.400 & 1.000 & 1.000 & 1.000 \\
0.450 & 1.000 & 1.000 & 1.000 \\
0.500 & 1.000 & 1.000 & 1.000 \\
\hline
\end{tabular}
\caption{results of young TPR for adversarial fair training on CelebA dataset under DI attack.}
\end{table*}

\begin{table*}[!h]
\centering
\begin{tabular}{llll}
\hline
M  &adv+pre & adv+in & adv+post\\
\hline
0.000 & 0.920 & 0.900 & 0.910 \\
0.050 & 0.890 & 0.880 & 0.880 \\
0.100 & 0.880 & 0.860 & 0.870 \\
0.150 & 0.860 & 0.850 & 0.860 \\
0.200 & 0.840 & 0.820 & 0.850 \\
0.250 & 0.810 & 0.810 & 0.810 \\
0.300 & 0.770 & 0.770 & 0.790 \\
0.350 & 0.740 & 0.700 & 0.740 \\
0.400 & 0.710 & 0.690 & 0.700 \\
0.450 & 0.690 & 0.660 & 0.690 \\
0.500 & 0.680 & 0.640 & 0.660 \\
\hline
\end{tabular}
\caption{results of young TNR for adversarial fair training on CelebA dataset under DI attack.}
\end{table*}

\begin{table*}[!h]
\centering
\begin{tabular}{llll}
\hline
M  &adv+pre & adv+in & adv+post\\
\hline
0.000 & 0.820 & 0.830 & 0.810 \\
0.050 & 0.800 & 0.810 & 0.800 \\
0.100 & 0.780 & 0.780 & 0.790 \\
0.150 & 0.760 & 0.770 & 0.780 \\
0.200 & 0.730 & 0.730 & 0.770 \\
0.250 & 0.700 & 0.700 & 0.720 \\
0.300 & 0.680 & 0.680 & 0.690 \\
0.350 & 0.660 & 0.660 & 0.660 \\
0.400 & 0.640 & 0.650 & 0.640 \\
0.450 & 0.610 & 0.630 & 0.620 \\
0.500 & 0.600 & 0.580 & 0.610 \\
\hline
\end{tabular}
\caption{results of elder TPR for adversarial fair training on CelebA dataset under DI attack.}
\end{table*}

\begin{table*}[!h]
\centering
\begin{tabular}{llll}
\hline
M  &adv+pre & adv+in & adv+post\\
\hline
0.000 & 0.900 & 0.920 & 0.910 \\
0.050 & 0.920 & 0.940 & 0.940 \\
0.100 & 0.940 & 0.970 & 0.970 \\
0.150 & 0.950 & 0.990 & 0.990 \\
0.200 & 0.970 & 1.000 & 1.000 \\
0.250 & 1.000 & 1.000 & 1.000 \\
0.300 & 1.000 & 1.000 & 1.000 \\
0.350 & 1.000 & 1.000 & 1.000 \\
0.400 & 1.000 & 1.000 & 1.000 \\
0.450 & 1.000 & 1.000 & 1.000 \\
0.500 & 1.000 & 1.000 & 1.000 \\
\hline
\end{tabular}
\caption{results of elder TNR for adversarial fair training on CelebA dataset under DI attack.}
\label{otnr-celeba}
\end{table*}

\end{document}